\PassOptionsToPackage{dvipsnames,table}{xcolor}
\documentclass[10pt,twocolumn,letterpaper]{article}

\usepackage{iccv}              % To produce the CAMERA-READY version
\definecolor{iccvblue}{rgb}{0.21,0.49,0.74}
\usepackage[pagebackref,breaklinks,colorlinks,allcolors=iccvblue]{hyperref}

\usepackage{graphicx}
\usepackage{wrapfig}
\usepackage{array}
\usepackage{bbding}
\usepackage{multirow}
\newcommand{\name}{\emph{ReME}}
\usepackage{colortbl}
\usepackage{soul}
\usepackage{algorithm}
\setuldepth{Berlin}
\usepackage{pifont}

\usepackage{amsmath}
\usepackage{fontawesome}
\usepackage{bm}
\usepackage[most]{tcolorbox}
\newtcbox{\myboxx}[1][gray]{on line, boxsep=2.5pt, boxrule=0pt, left=-0.5pt,right=-0.5pt,top=-1pt,bottom=-1pt, colback=gray, colframe=gray,arc=1.4mm}
\definecolor{customPurple}{HTML}{DB62B5}
\def\paperID{6604} % *** Enter the Paper ID here
\def\confName{ICCV}
\def\confYear{2025}

\title{ReME: A Data-Centric Framework\\ for Training-Free Open-Vocabulary Segmentation}
\author{Xiwei Xuan, Ziquan Deng, and Kwan-Liu Ma\\
University of California, Davis\\
{\tt\small \{xwxuan, ziqdeng, klma\}@ucdavis.edu}
}

\begin{document}
%%%%%%%%%%%%%%%%%%%%%%%%%%%%%
\maketitle
\begin{abstract}
Training-free open-vocabulary semantic segmentation (OVS) aims to segment images given a set of arbitrary textual categories without costly model fine-tuning. Existing solutions often explore attention mechanisms of pre-trained models, such as CLIP, or generate synthetic data and design complex retrieval processes to perform OVS. However, their performance is limited by the capability of reliant models or the suboptimal quality of reference sets. In this work, we investigate the largely overlooked data quality problem for this challenging dense scene understanding task, and identify that a high-quality reference set can significantly benefit training-free OVS. With this observation, we introduce a data-quality-oriented framework, comprising a data pipeline to construct a reference set with well-paired segment-text embeddings and a simple similarity-based retrieval to unveil the essential effect of data. Remarkably, extensive evaluations on ten benchmark datasets demonstrate that our method outperforms all existing training-free OVS approaches, highlighting the importance of data-centric design for advancing OVS without training.
Our code is available \href{https://github.com/xiweix/ReME}{here}.

\end{abstract}
    
\section{Introduction}
\label{sec:intro}

\begin{figure}[t]
  \centering
   \includegraphics[width=1\linewidth]{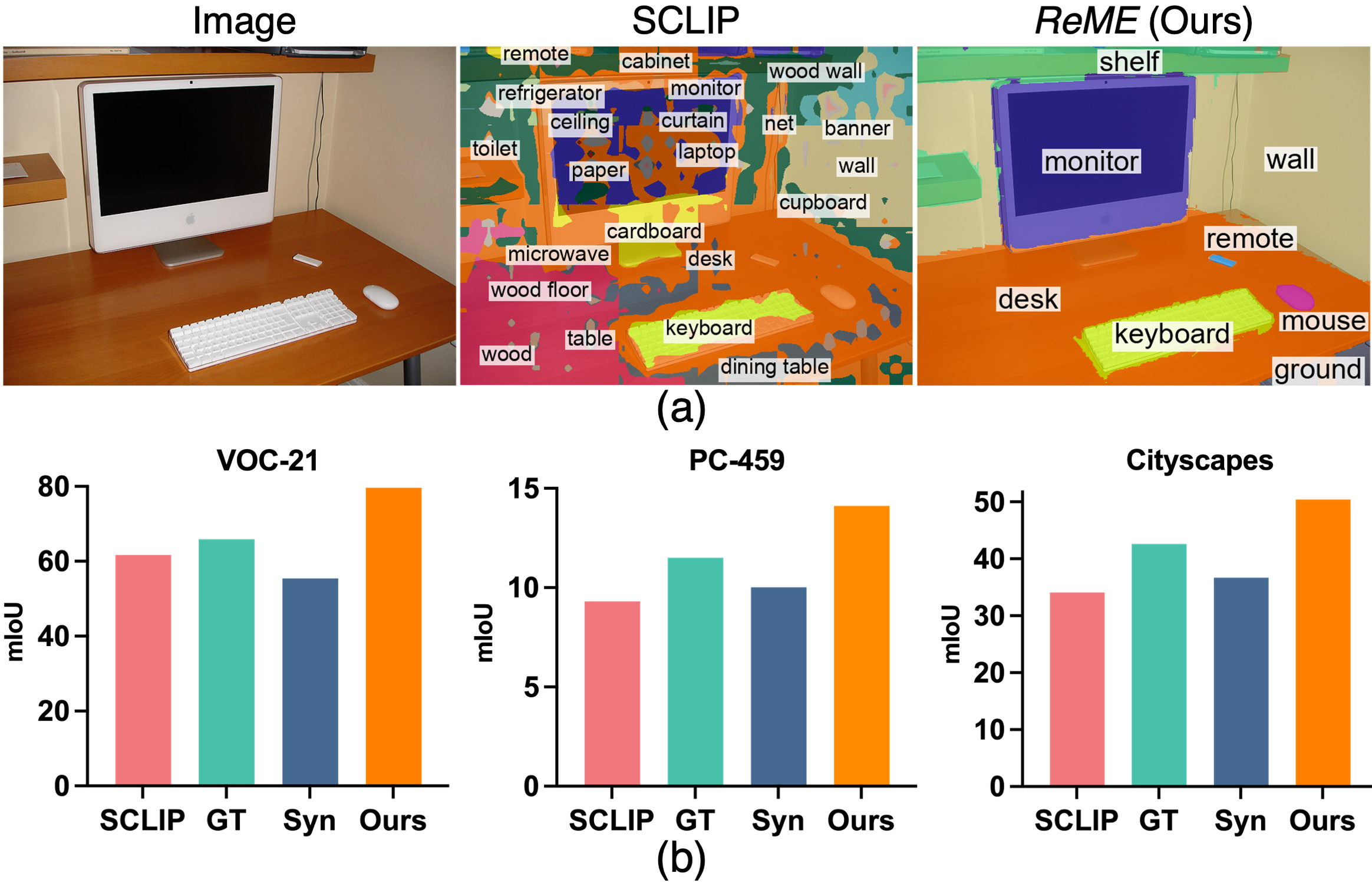}
   \caption{\textbf{(a)} Qualitatively compare \name~with SCLIP~\cite{wang2024sclip} (a representative CLIP-attention-based method) on an example from ADE20K~\cite{zhou2019semantic} with $150$ classes. Both methods have no mask post-processing. \name~produces higher-quality masks with a simple segmentation algorithm, yet SCLIP generates noisy masks paired with many irrelevant classes, such as \textit{``curtain''} and \textit{``microwave''}, etc. \textbf{(b)} Performance comparison between SCLIP and retrieval-based OVS using different reference sets. \textcolor[HTML]{6BBDAB}{\textbf{GT}}: ground-truth segment-label of COCO Stuff~\cite{caesar2018coco}; \textcolor[HTML]{4D688D}{\textbf{Syn}}: synthetic reference set from FreeDA~\cite{barsellotti2024training};  \textcolor{orange}{\textbf{Ours}}: our reference set from real images with quality refinement.
   }
   \label{fig:mask_compare_main}
\end{figure}

Open-vocabulary semantic segmentation (OVS) aims to segment images according to a set of arbitrary textual categories.
% , enabling flexible and adaptable scene understanding. 
While conventional approaches rely on training or fine-tuning large models with segment-text~\cite{xu2023side,ding2022decoupling,ghiasi2022scaling,liang2023open,yu2023fcclip,luo2023segclip,mukhoti2023open,cho2024cat} or image-text~\cite{xu2022groupvit,yi2023simple,cha2023learning,xing2024rewrite,chen2023exploring,xu2023learning,wu2024image,wang2024sam} datasets, these methods incur significant annotation and computational costs. Recent advancements in vision-language models (VLMs), such as CLIP~\cite{radford2021learning}, have spurred interest in training-free OVS methods that leverage these pre-trained models without additional fine-tuning.

Some approaches assume VLMs have sufficient classification capabilities to recognize semantics, and attempt to enhance their pixel-level localization by modifying the attention mechanisms~\cite{zhou2022extract,shao2024explore,sun2024clip,wang2023diffusion,lan2024proxyclip,li2023tagclip,wang2024sclip}. However, VLMs are trained under weak image-text supervision~\cite{li2023clip}, and merely tweaking their attention modules post-hoc cannot break through the inherent ceiling imposed by coarse training signals~\cite{guo2023robustifying}. As a result, such approaches often struggle with low-quality segmentation masks and noisy labels. As shown by an example in Fig.~\ref{fig:mask_compare_main} (a), methods relying on CLIP attention often fail to even match the segmentation quality of an algorithm-based segmenter~\cite{felzenszwalb2004efficient}, which, despite being class-agnostic, can propose perceptually meaningful image regions. (Refer to Fig.~\ref{fig:data_more_qua} for more examples).

Confronted with such limitations of VLMs, the retrieval paradigm offers a key inspiration: correcting the vulnerabilities of VLMs by reference substances in an external knowledge base. We investigate this from the data-centric perspective -- assuming there exists a high-quality reference set for OVS, can we unlock the potential of retrieval for better results? Specifically, we experiment with ground truth (GT) segment-text pairs from COCO Stuff~\cite{caesar2018coco}, viewing it as a ``high-quality'' reference set. Using a class-agnostic segmentation algorithm~\cite{felzenszwalb2004efficient} for test images, we apply a simple similarity-based strategy (refer to Sec.~\ref{subsec:ref_retrieval}) for retrieving and aggregating mask labels. No models or complex retrieving algorithms are involved to unveil the fundamental data capabilities. As shown in Fig.~\ref{fig:mask_compare_main} (b), retrieval from GT annotations consistently surpasses SCLIP~\cite{wang2024sclip}, which is a well-performed method using CLIP attention. This observation indicates the significant potential of data for enhancing OVS with retrieval.

However, in the absence of labor-intensive GT annotations, despite the anticipated performance gain, existing retrieval-based methods remain modest compared to attention-based approaches. To fill the data gap, they often resort to diffusion models (DMs)~\cite{rombach2022high} to generate synthetic images, utilizing text-corresponding attention masks to construct reference sets~\cite{karazija2024,barsellotti2024training,barsellotti2024fossil,wang2024image}. However, synthetic data often lacks the realism and richness of real images, making them a suboptimal retrieval resource~\cite{fan2024scaling,singh2024synthetic,xuan2024attributionscanner}. Fig.~\ref{fig:mask_compare_main} (b) compares retrieval results from the COCO Stuff GT and a state-of-the-art synthetic reference set from FreeDA~\cite{barsellotti2024training}, constructed from COCO captions. The comparison indicates that the synthetic reference set falls significantly short of its GT counterpart, highlighting its limitations compared to real images. Some methods employ a quick fix of data bottleneck by designing sophisticated retrieval strategies to tolerate suboptimal data~\cite{wang2024image}, which is less preferred as the central issue remains unaddressed~\cite{xuan2024slim}. 

In this work, we aim to construct \textbf{a well-aligned, rich, and contextually relevant set of segment-text pairs from real images}, investigating the capabilities of such data to benefit training-free OVS. Specifically, we introduce \name, a data-centric framework that \underline{Re}fines \underline{M}ulti-modal \underline{E}mbeddings for retrieval-based, training-free OVS.

\name~harnesses the intrinsic capabilities of VLMs with the aggregation effect -- while they may not yield satisfactory results for individual instances, VLMs demonstrate a certain degree of effectiveness when analyzing data at a collective level~\cite{schuhmann2021laion,schuhmann2022laionb,radenovic2023filtering,nguyen2024improving,zhu2023chatgpt,zhang2023labelvizier,deng2024reliable}. Using only images as input, the data pipeline of \name~starts by constructing a semantically rich base set with VLM-paired image segments and textual labels. With the observation of superior discriminativeness in same-modal data features, \name~leverages their collective patterns to recalibrate cross-modal misalignments introduced by the VLM, producing a reference set with enhanced quality. In the inference, we apply a basic class-agnostic segmentation algorithm to test images, then align and aggregate the proposed masks into final predictions by referring to our reference set with a simple similarity-based strategy.

As shown in Fig.~\ref{fig:mask_compare_main} (b), compared to using GT annotations for reference, our reference set indicates even better OVS capacity. Specifically, governed by our data quality consideration in semantic richness and alignment correctness, \name~brings significantly further performance gains for challenging benchmarks, such as PC-459~\cite{mottaghi2014role}, and domain-specific scenarios, such as Cityscapes~\cite{cordts2016cityscapes}.

Extensive experiments on ten OVS benchmarks demonstrate the superior performance of \name~over existing training-free methods.
Overall, our main contributions are summarized as follows:
\begin{itemize}
    \item We introduce \name, a data-centric framework for training-free OVS by constructing high-quality segment-text data from real images, with a simple retrieval process to perform OVS without any model fine-tuning.
    \item We identify the critical yet underexplored role of data quality, and provide a data pipeline for refining multi-modal data collectively from an intra-modal perspective, which unveils the data potential to regularize or complement pre-trained models.
    \item Extensive experiments on ten  benchmark datasets demonstrate the consistent performance gains of \name~over $14$ training-free OVS baselines.
\end{itemize}

\section{Related Work}
\label{sec:related_work}

\begin{figure*}[th]
\setlength{\belowcaptionskip}{-0.5cm}
  \centering
   \includegraphics[width=1\linewidth]{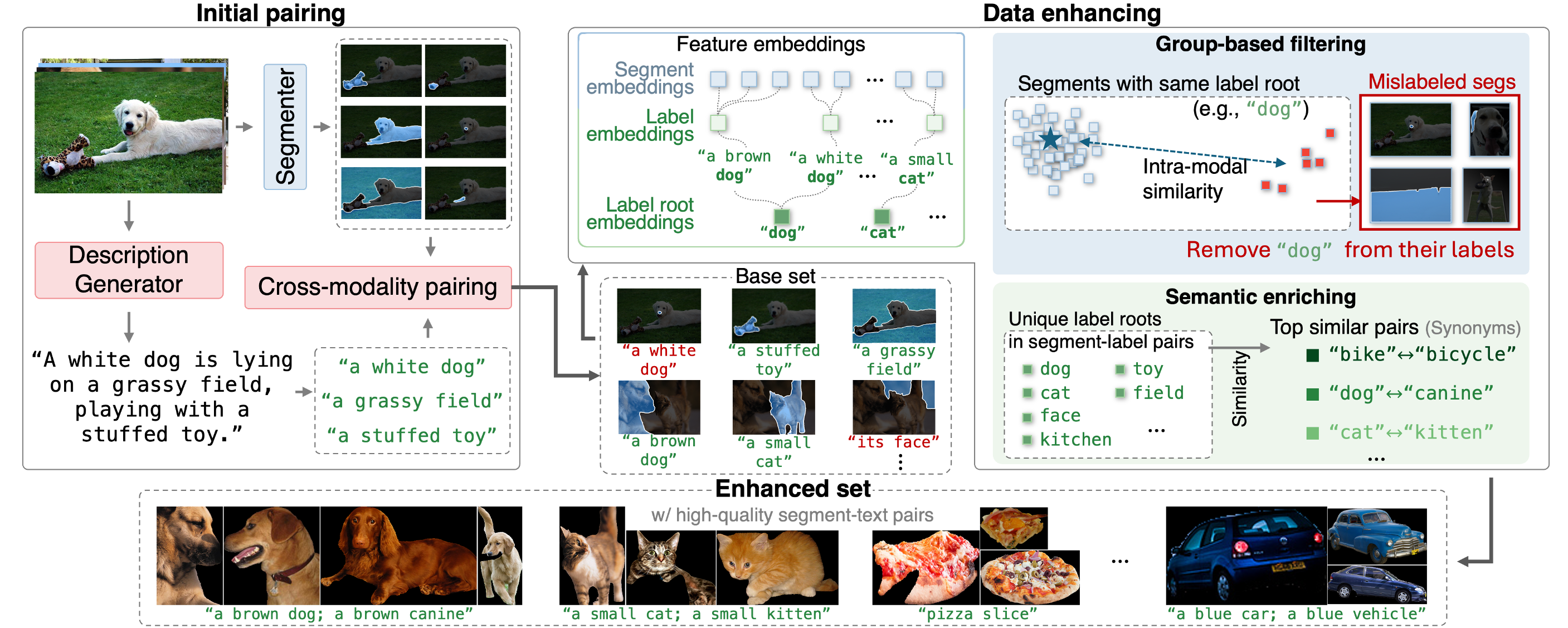}
   \caption{\textbf{The data pipeline of \name~to construct a high-quality reference set.} The pipeline includes two phases: the \textit{\textbf{initial pairing}} that produces pre-matched segment-text pairs using images as input; and the \textit{\textbf{data enhancing}} that performs group-based filtering according to more discriminative intra-modal similarity among segment embeddings and semantic enriching with similar labels.}
   \label{fig:data_pipeline}
\end{figure*}

\noindent \textbf{Training-Free OVS.}
Many training-free OVS methods analyze attention mechanisms~\cite{vaswani2017attention,xuan2024suny,xuan2022vac} of VLMs or diffusion models to produce segmentation results~\cite{zhou2022extract,lan2024proxyclip,wang2023diffusion,shao2024explore,rewatbowornwong2023zero,bousselham2024grounding,wysoczanska2024clip,luo2024emergent}.
However, their reliant models trained without pixel-level supervision may produce inferior results, constraining the upper bound of these approaches. To improve the situation, methods such as CaR~\cite{sun2024clip} leverage multiple CLIP models to iteratively refine segmentation results, which are less practical due to significant inference costs.
Another line of methods~\cite{shin2022reco,karazija2024,barsellotti2024training,barsellotti2024fossil,gui2024knnclip,wang2024image} constructs a reference set with extensive segment-text pairs, which serves as reference knowledge to support better alignment between test segments and text.
Most of such methods either use synthetic data~\cite{karazija2024,barsellotti2024training,barsellotti2024fossil,wang2024image} that falls short of diversity and realism~\cite{fan2024scaling,singh2024synthetic}, or opt for sophisticated retrieval strategy or shortcut like adding GT classes to bypass the data quality issues~\cite{wang2024image,nguyen2024dataset}. To fill the gap, our framework investigates the importance of data for this challenging task.

\noindent \textbf{Multi-Modal Datasets Curation.}
High-quality datasets have driven remarkable progress in multi-modal models~\cite{wang2024use,fang2024data,mahmoud2024sieve, cha2023learning,peng2023kosmos,xiao2024florence,xuan2025vista}. Recent techniques of image-text data curation typically involve two branches: \textit{data filtering}~\cite{cao2023less,gadre2024datacomp} that removes noisy data, and \textit{data improvement}~\cite{lai2025veclip,zhu2023chatgpt} that refines multi-modal alignment.
Despite training better VLMs is often the objective of data curation, these models themselves are also used for curating better data. For instance, CLIP is frequently used for data filtering with CLIP score (i.e., similarity between visual and text features)~\cite{schuhmann2021laion,schuhmann2022laionb,radenovic2023filtering}, and some methods use BLIP-2 to reduce semantic noise~\cite{nguyen2024improving,zhu2023chatgpt}. As both \textit{data filtering} and \textit{data improvement} address essential aspects of data quality, we consider both of them in our approach for more comprehensive data enhancement. 
In addition, multi-modality data curation in fine-grain such as segment-text pairs remains underexplored, requiring further research to support more precise, context-aware applications.
Beyond OVS, our proposed data cleaning methodology has the potential for broader applicability across other contexts.

\noindent \textbf{Pre-Training Models.} 
Pre-training models often have high reusability and adaptability, which can be easily reused in various approaches without further fine-tuning~\cite{shin2022reco,luo2024emergent,karazija2024,zhang2024corrclip,yan2025vislix}. For instance, \textit{Vision-Language Models (VLMs)} like CLIP~\cite{radford2021learning} and ALIGN~\cite{jia2021scaling} link text with images for zero-shot image classification. \textit{Multimodal Large Language Models (MLLMs)}, such as LLaVA~\cite{liu2023llava}, BLIP-2~\cite{li2023blip}, and GPT-Vision~\cite{achiam2023gpt}, support diverse tasks such as captioning, reasoning, and visual chat. Vision models like DINO~\cite{caron2021emerging} and DINOv2~\cite{oquab2024dinov} provide universal visual feature representations. Despite their success, such models are not almighty solutions, having various limitations for specified tasks~\cite{lai2024lisa}. In this work, we study how to effectively harness the value of data to compensate for the shortcomings of such models in a training-free manner.

\section{Our Approach}
\label{sec:method}

The goal of an OVS method is to segment an input image given an arbitrary set of textual labels. Our training-free framework \name~consists of a data pipeline with two key phases, \textbf{\textit{initial pairing}} and \textbf{\textit{data enhancing}}, to construct a high-quality reference set of segment-text pairs (Fig.~\ref{fig:data_pipeline}); followed by a simple \textbf{\textit{similarity based retrieval}} process to unveil the performance gains enabled by our data (Fig.~\ref{fig:retrieval}).

\subsection{Initial Pairing}
\label{subsec:ref_construction}
Our data pipeline starts by constructing a diverse base set with segment-text pairs, where we aim to ensure broad coverage of various objects and contextual elements.

As shown in Fig.~\ref{fig:data_pipeline}, for each input image, a segmenter generates class-agnostic segment masks, and an image description generator produces semantically rich descriptions. Following~\cite{peng2023kosmos,xiao2024florence}, we extract noun phrases (i.e., nouns with descriptive modifiers such as \textit{``a lovely white rabbit''}) from the descriptions to produce candidate textual labels, which reserve richer semantics than pure nouns. Then, we pair the segments and labels using a capable VLM.
Despite the noises, CLIP still provides a certain degree of correctness for segment labeling, as evidenced by CLIP-based OVS methods~\cite{zhou2022extract,shao2024explore,sun2024clip,lan2024proxyclip,gui2024knnclip}.
We thus leverage CLIP to perform initial pairing. Given an image, we extract CLIP embeddings of its segments and noun phrases, \( \bm{S} \in \mathbb{R}^{m \times d} \) and \( \bm{L} \in \mathbb{R}^{n \times d} \), respectively, where \( m \), \( n \) is the number of segments and labels, and \( d \) is the embedding dimension. Note that all embeddings across the paper are L2 normalized so that \(X \cdot Y^T\) is equivalent to their cosine similarity \(\langle X, Y\rangle\).
We then compute \( sim = \bm{S} \cdot \bm{L}^T \in \mathbb{R}^{m \times n} \) to pair each label with its closest segment, where \( {sim}_{ij} \) is the similarity between \( \bm{S}_i \) and \( \bm{L}_j \). The top-matching \( S_{i^*} \) of \( L_j \) is identified by: \(i^* = \arg\max_{i} {sim}_{ij}\). The unpaired segments are dropped to reduce the redundancy in class-agnostic image segments.

Despite including diverse segments paired with noun phrase labels, the base set still has unaddressed quality concerns, which is our major focus in the following phase.

\begin{figure}[tb!]
\setlength{\belowcaptionskip}{-0.5cm}
  \centering
   \includegraphics[width=1\linewidth]{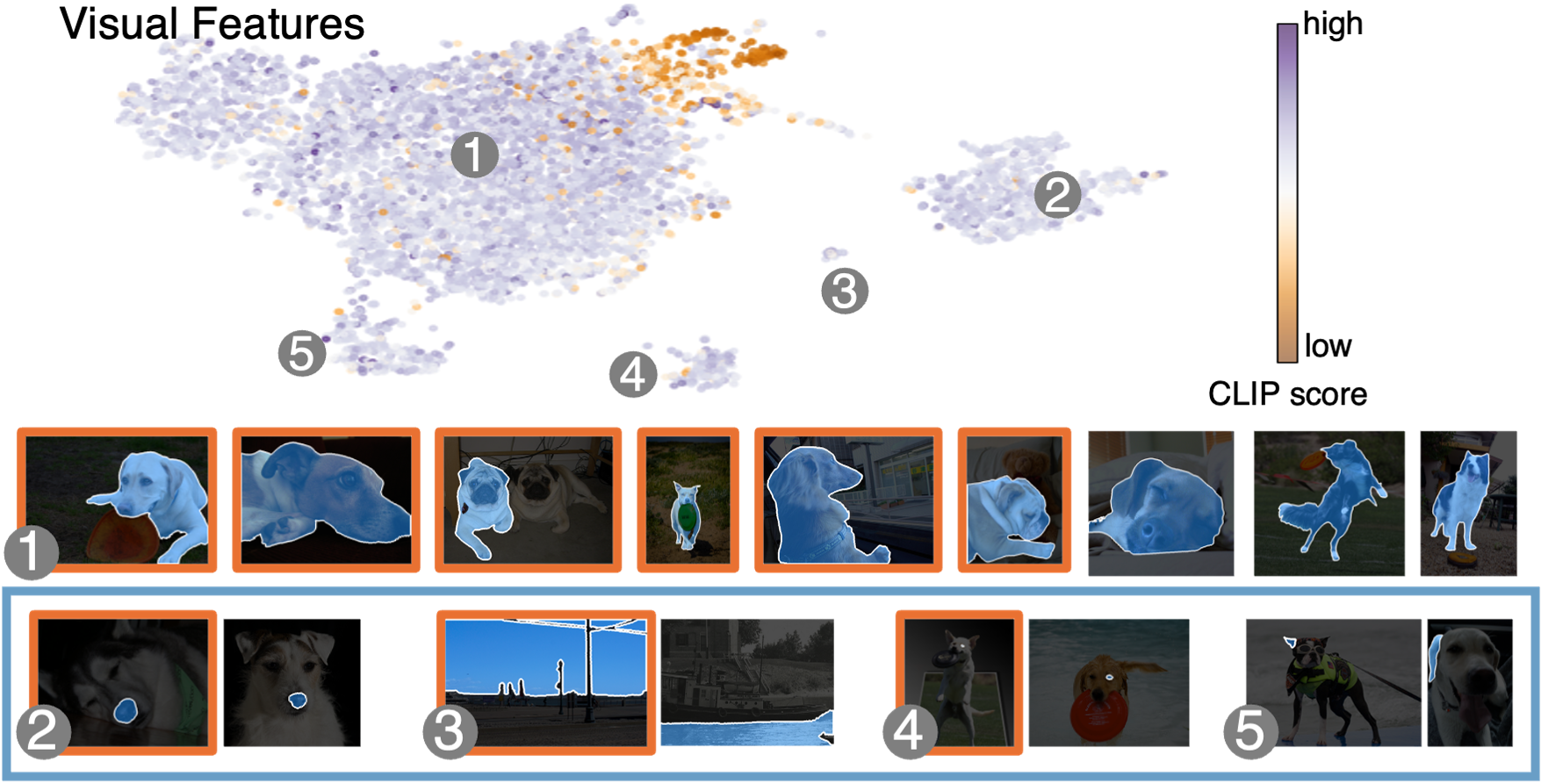}
   \caption{\textbf{The superiority of intra-modality over cross-modality for data issue detection.} The plot provides the UMAP projection of segment embeddings in the base set labeled as ``dog'', colored by cross-modal similarity scores (CLIP scores). {\textcolor[HTML]{759DC2}{\textbf{Blue boxes}}} highlight misalignments detected by our filtering; {\textcolor[HTML]{D97843}{\textbf{orange boxes}}} are those detected by low CLIP scores, which remove correct pairings while leaving many misalignments unaddressed.}
   \label{fig:seg_rep_space}
\end{figure}

\subsection{Data Enhancing}
\label{subsec:refset_refinement}
As shown in Fig.~\ref{fig:data_pipeline}, through initial pairing, we obtain a base set with paired segments and labels.
Various factors negatively affect its quality: (1) irrelevant textual labels caused by object hallucinations of the description generator, (2) meaningless or partial segments from over-segmentation, and (3) segment-text pairing errors. These challenges are essentially introduced by the ambiguity in cross-modality. We opt for navigating them with intra-modality operations---which we find to be more discriminative to effectively clean and enrich such data.

Unlike the common practice of dropping segment-label pairs with \ul{low cross-modal similarity scores (i.e., CLIP scores)}~\cite{schuhmann2021laion,schuhmann2022laionb,radenovic2023filtering}, we reexamine data features and test whether CLIP scores can favorably clean noisy data. As shown in Fig.~\ref{fig:seg_rep_space}, our observation brings two key insights: (1) CLIP scores tend to mistakenly remove correct pairs while leaving many misalignments unaddressed; (2) Visual features, i.e., embeddings of image segments, have favorable discriminative power for data issue detection. (Please refer to Sec.~\ref{subsec:ablation} for extensive discussions.)

\noindent\textbf{Group-Based Filtering.}
Guided by our observation, in this phase, we detect misalignments by leveraging the discriminativeness of intra-modal features. Specifically, we aim to leverage the visual features of segments to identify outliers in their own modality. First, for our labels in the format of noun phrases, we view those with the same ``root'' noun as an identical label. Next, we group each single segment-label pair according to identical labels. For example, segments labeled by \textit{``a small dog''} or \textit{``an adorable dog''} belong to the same group denoted by the root \textit{``dog.''} Note that we reserve all descriptive modifiers in the phrases to keep semantic richness; ``roots'' are just used for grouping.

This is an important step for detecting data misalignments. Since each group of segments corresponds to an identical label, their visual features should be inherently consistent, and misaligned data would be automatically highlighted as outliers. As shown in Fig.~\ref{fig:seg_rep_space}, the majority of segments marked by \myboxx{\textcolor{white}{1}} accurately depict dogs, while others are misalignments located as outliers in the projection, indicating their distinctive features.

In detail, we define the segment center \(S_{center}\) of each group, computed by the median of all visual features. \(S_{center}\) denotes the representative features for correctly labeled segments. We then compute the intra-modal similarity between each segment in the current group and \(S_{center}\), indicating the possibility of a segment being mistakenly labeled. \(\delta_{filter}\) percent of segments with the lowest scores are considered mistakenly paired, where we remove the corresponding noun phrases from their labels. By performing this filtering guided by intra-modality feature similarity, we effectively reduce misalignment in the base set.

\noindent\textbf{Semantic Enriching.}
By clearing outliers in each group, we have significantly mitigated segment-text misalignments. However, data enhancement remains incomplete in terms of semantic diversity -- While statistical analysis of the entire label corpus of our base set reveals the textual modality includes different words for the same concept (e.g., \textit{``bike''} and \textit{``bicycle''}), such alternatives are missing in individual segments. This results in a lack of semantic richness, as we expect our method to recognize concepts in various representations. Consistently, we design an intra-modality-based approach to diversify our labels.
As shown in Fig.~\ref{fig:data_pipeline}, we collect embeddings of identical labels, i.e., the ``root'' noun of our noun phrase labels, denoted as \( \{L_1, \dots, L_n \}\). Then, we compute their pair-wise cosine similarity \(\langle L_i, l_j \rangle\), and identify \(k\) top-similar pairs to be treated as synonyms. Next, label semantics are enriched by adding these synonyms. For instance, \textit{``cat''}-\textit{``kitten''} is a pair of synonyms, so, if a segment has a label \textit{``a small cat''}, we add \textit{``a small kitten''} to become its label as well. Those synonym-enriched noun phrases are our finalized labels. This process further diversifies our textual descriptions by leveraging the global semantic richness present within the data. Note that, an LLM could produce even better semantic enriching, while we opt for a lightweight-yet-effective solution to avoid the additional overhead of large models.

\subsection{Similarity-Based Retrieval}
\label{subsec:ref_retrieval}

Following Sec.~\ref{subsec:refset_refinement}, we have constructed a reference set with well-paired segments and labels and stored their embeddings.
At inference time, given a test image and a set of target classes, our goal is to assign pixel-level labels to the test image by referring to the reference set.
Inspired by Tip-Adapter~\cite{zhang2022tip}, we perform a simple retrieval strategy grounded in feature similarity, where labels of test segments are estimated by their closest matches in the reference set.

\begin{figure}[tb]
  \centering
   \includegraphics[width=1\linewidth]{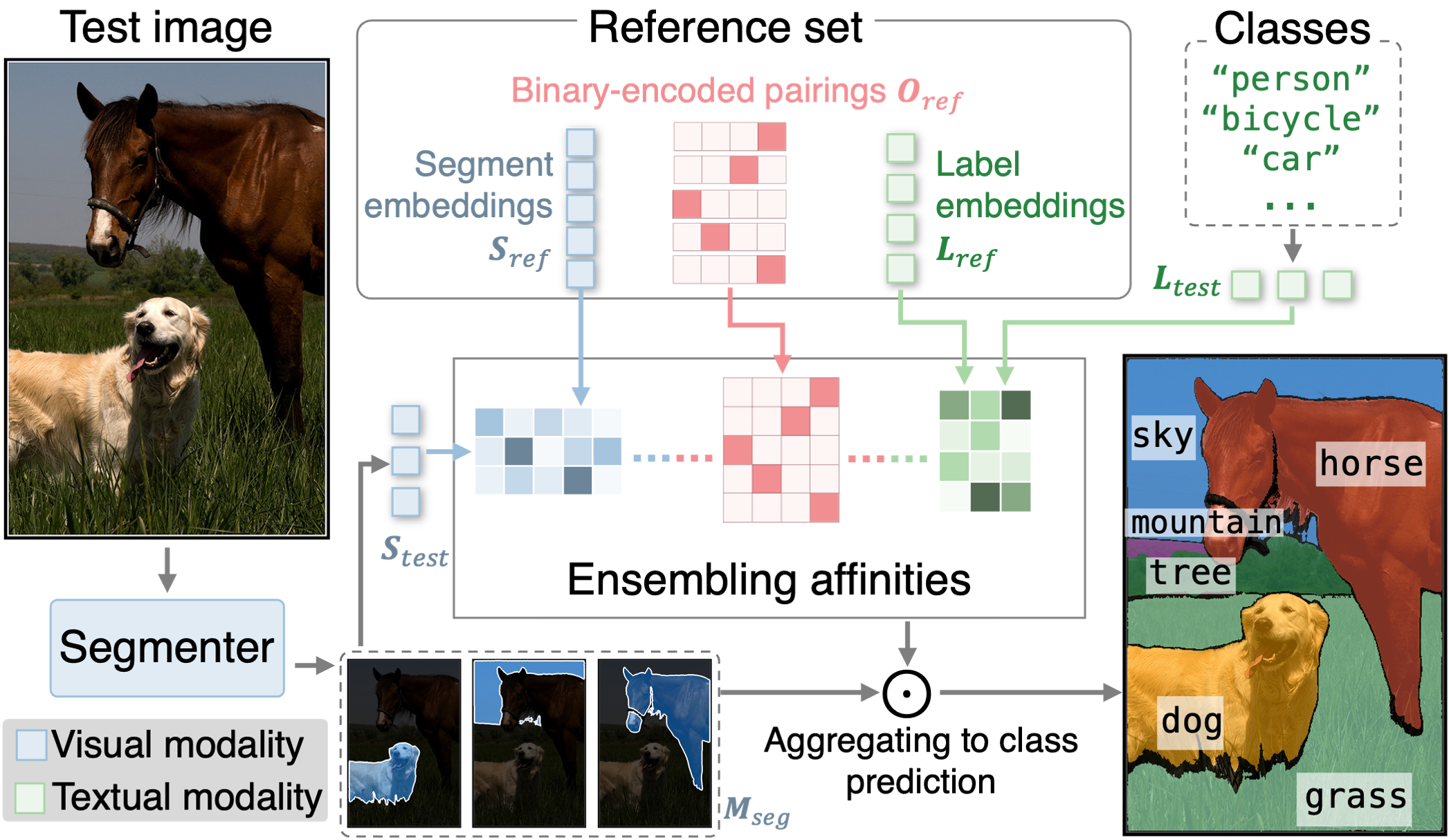}
   \caption{The process of similarity-based retrieval from the reference set given a test image and classes.
   Note that the same modality embeddings (i.e., reference \& test segments; reference \& test labels) are encoded by the same feature encoder, respectively.
   }
   \label{fig:retrieval}
\end{figure}

As shown in Fig.~\ref{fig:retrieval}, the reference set includes embeddings of \(m\) segments \(\bm{S}_{\text{ref}} \in \mathbb{R}^{m \times d_1}\) and \(n\) labels \(\bm{L}_{\text{ref}} \in \mathbb{R}^{n \times d_2}\). The label assignments are encoded by a binary matrix \(\bm{O}_{\text{ref}} \in \mathbb{R}^{m \times n}\), with an entry $0$ indicating the corresponding label is absent in the segment and $1$ otherwise.

Given a test image, we segment it into $k$ class-agnostic masks, denoted by \(\bm{M}_{\text{seg}} \in \mathbb{R}^{k \times h \times w}\), where \(h \times w\) is the image size. We extract the segment and test class embeddings \(\bm{S}_{\text{test}} \in \mathbb{R}^{k \times d_1}\) and \(\bm{L}_{\text{test}} \in \mathbb{R}^{c \times d_2}\), using the same encoders applied to the reference set, where \(d_1\) and \(d_2\) are the embedding dimensions of the image and text modalities, respectively.
The affinity between test segments and reference set labels is computed as:
\begin{equation}
A_1 = \text{Softmax}(\bm{S}_{\text{test}} \cdot \bm{S}_{\text{ref}}^T) \cdot \bm{O}_{\text{ref}},
\label{eqn:affinity_test_seg_label}
\end{equation}
where \(A_1 \in \mathbb{R}^{k \times n}\), and Softmax normalizes the affinities to formulate the probability distribution of labels for each test segment. 
Similarly, the affinity matrix between the reference labels and test classes is represented as:
\begin{equation}
A_2 = \text{Softmax}(\bm{L}_{\text{ref}} \cdot \bm{L}_{\text{test}}^T),
\label{eqn:affinity_label_class}
\end{equation}
where \(A_2 \in \mathbb{R}^{n \times c}\). Ensembling the affinities, for $k$ test segments, the predicted logit is defined as: \(\bm{P}_{\text{seg}} = A_1 \cdot A_2, \bm{P}_{\text{seg}} \in \mathbb{R}^{k \times c}.\)
Then, we calculate pixel-wise label probabilities \(\bm{P}_{\text{test}} \in \mathbb{R}^{h \times w \times c}\) by aggregating ensembled affinities according to the segment masks \(\bm{M}_{seg}\), where the probability of class \(j\) at coordinates \((x, y)\) is:
\begin{equation}
\bm{P}_{\text{test}}^{(x, y, j)} = \sum_{i=1}^{k} \bm{P}_{\text{seg}}^{(i, j)} \cdot \bm{M}_{\text{seg}}^{(i, x, y)}.
\label{eqn:logit}
\end{equation}
Lastly, the predicted mask \(\hat{l} \in \mathbb{R}^{h \times w}\) is determined by:
\(
\hat{l}^{(x, y)} = \arg\max_{j} \bm{P}_{\text{test}}^{(x, y, j)},
\)
where \(\hat{l}^{(x, y)} \in [0, c-1]\) indicates class predictions, which conclude our retrieval phase.

\section{Experiments}
\label{sec:experiments}
\begin{figure*}[th]
  \centering
   \includegraphics[width=1\linewidth]{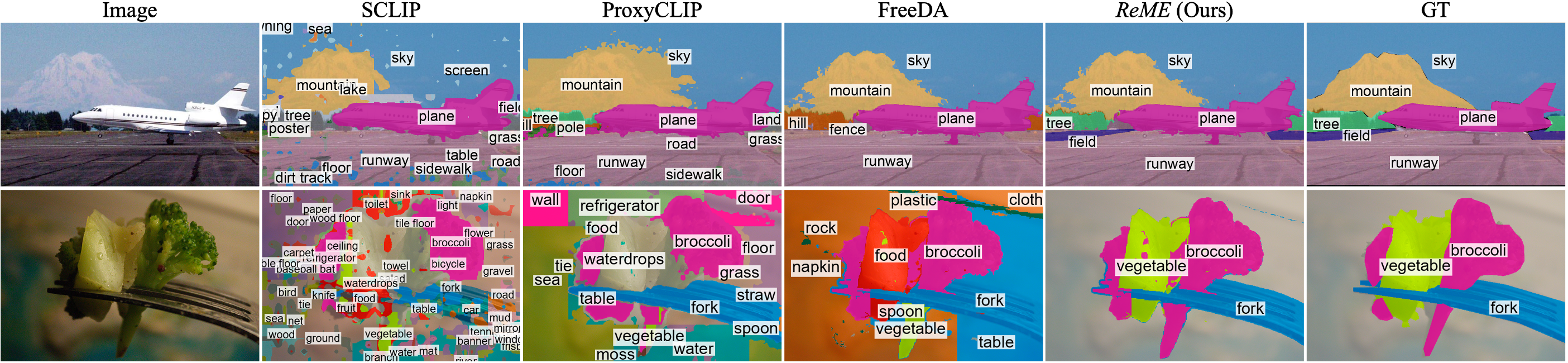}
   \caption{\textbf{Qualitative results of \name~in comparison with other training-free OVS methods.} Examples are from ADE20K~\cite{zhou2019semantic} (w/ 150 classes) and COCO Stuff~\cite{caesar2018coco} (w/ 171 classes), respectively. SCLIP is based on CLIP attention; ProxyCLIP enhances CLIP attention with DINO features; FreeDA and \name~are retrieval-based methods, adopting the same superpixel-algorithm~\cite{felzenszwalb2004efficient} for class-agnostic segmentation. We observe increasing quality of OVS results from left to right, with less noise in both masks and assigned labels.}
   \label{fig:data_more_qua}
\end{figure*}

\begin{table*}[tbh]
\centering
\resizebox{\textwidth}{!}{%
\begin{tabular}{l c c c c c c c c c c c}
\toprule[0.8pt]
\multirow{2}{*}{Methods}    & \multirow{2}{*}{Post-processing}       & \multicolumn{10}{c}{mIoU}                                \\ \cline{3-12} 
 &  & VOC-20 & VOC-21 & City & PC-59 & PC-60 & Object & Stuff & A-150      & PC-459  & A-847  \\ \hline
\multicolumn{3}{l}{\textit{Training-free Methods without SAM}}            &        &            &       &       &             &            &              &         &         \\
GEM~\cite{bousselham2024grounding}  &\ding{55} &46.2 &24.7 &-&32.6 &21.2 &- &15.1 &10.1 &4.6 &3.7\\
MaskCLIP~\cite{zhou2022extract}    &  \Checkmark    & 74.9   & 38.8   & 12.6          & 25.5  & 23.6  & 20.6        & 14.6       & 9.8          & -       & -         \\
ReCo~\cite{shin2022reco}           &  \Checkmark                                      & 62.4   & 27.2   & 23.2       & 24.7  & 21.9  & 17.3        & 16.3       & 12.4         & -       & -       \\
SCLIP~\cite{wang2024sclip} & \Checkmark &83.5 &61.7 &34.1 &36.1 &31.5 &32.1 &23.9 &17.8 &9.3 &6.1\\
CaR~\cite{sun2024clip}  & \Checkmark          & \underline{91.4}   & 67.6     & 15.1            & 39.5  & 30.5  & 36.6 & 11.2 & 17.7         & 11.5     & 5.0 \\
NACLIP~\cite{hajimiri2024pay} &\Checkmark &83.0 &64.1 &38.3 &38.4 &35.0 &36.2 &25.7 &19.1 &9.0 &6.5\\
CLIPtrase~\cite{shao2024explore}     &\Checkmark           & 81.2   & 53.0  & 21.1      & 34.9 & 30.8 & \underline{39.6}       & 24.1      & 17.0        & 9.9    & 5.9    \\
PnP~\cite{luo2024emergent}       &   \Checkmark      & 79.1  & 51.3   & 19.3      & 31.0    & 28.0    & 36.2        & 17.9       & 14.2         & 5.5    & 4.2    \\
FreeDA~\cite{barsellotti2024training}     & \Checkmark  & 87.9   & 55.4   & 36.7       & \underline{43.5}  & \underline{38.3}  & 37.4        &\underline{28.8}       & 22.4         & 10.2  & 5.3 \\
ProxyCLIP~\cite{lan2024proxyclip}        & \ding{55}    & 83.2   &  60.6    & \underline{40.1}         & 37.7  & 34.5  & 39.2      &  25.6      & \underline{22.6}         &  11.2  & \underline{6.7}   \\
DiffSegmenter~\cite{wang2023diffusion}      &  \Checkmark      &  71.4     & 60.1   & -         & 27.5  &  25.1     & 37.9        &   -         & -             &  -       & -        \\
OVDiff~\cite{karazija2024}     & \Checkmark   & 80.9   & \underline{68.4}   & 23.4       & 32.9  & 31.2  & 36.2        & 20.3       & 14.1         & \underline{12.0}       & 6.6       \\
\rowcolor[HTML]{EFEFEF} 
\textbf{\name~(Ours)}  & \ding{55}  & \textbf{92.3}       &\textbf{79.6}      & \textbf{50.4}           & \textbf{44.9}     &\textbf{41.6}        & \textbf{45.5}            & \textbf{33.1}           & \textbf{26.1}             &\textbf{14.1}        & \textbf{8.4} \\
 \hline
\rowcolor[HTML]{EFEFEF} 
\textcolor{gray}{\textbf{\name~(Ours - VOC)}}  & \textcolor{gray}{\ding{55}}  & \textcolor{gray}{84.7}       &\textcolor{gray}{\textbf{75.0}}      & \textcolor{gray}{\textbf{43.9}}           & \textcolor{gray}{40.9}     &\textcolor{gray}{\textbf{38.7}}        & \textcolor{gray}{\textbf{40.8}}            & \textcolor{gray}{22.6}           & \textcolor{gray}{\textbf{25.2}}             &\textcolor{gray}{12.8}       & \textcolor{gray}{\textbf{8.3}} \\
\rowcolor[HTML]{EFEFEF} 
\textcolor{gray}{\textbf{\name~(Ours - ADE)}}  & \textcolor{gray}{\ding{55}}  & \textcolor{gray}{84.3}       &\textcolor{gray}{\textbf{72.3}}      & \textcolor{gray}{\textbf{42.1}}           & \textcolor{gray}{\textbf{44.0}}     &\textcolor{gray}{\textbf{39.7}}        & \textcolor{gray}{35.8}            & \textcolor{gray}{27.0}           & \textcolor{gray}{\textbf{26.0}}             &\textcolor{gray}{\textbf{13.2}}       & \textcolor{gray}{\textbf{8.6}} \\

\hline
\multicolumn{3}{l}{\textit{Training-free Methods with SAM}}     &        &            &       &       &             &            &              &         &         \\
RIM~\cite{wang2024image}      & \ding{55}    & 77.8   & -      & -          & 34.3  & -     & \underline{44.5}        & -          & -            & -       & -       \\
CaR w/ SAM~\cite{sun2024clip} & \ding{55} &- &70.2 &16.9 &40.5 &31.1 &37.6 &12.4 &17.9 &\underline{11.8}   & 5.7\\
CLIPtrase w/ SAM~\cite{shao2024explore} & \ding{55} &82.3 &57.1 &- &36.4 &32.0 &44.2 &24.8 &17.2 &10.6 &\underline{6.0} \\
ProxyCLIP w/ SAM~\cite{lan2024proxyclip}  & \ding{55} &80.4 &59.3 &37.0 &37.0 &33.6 &35.4 &25.0 &\underline{19.1} &6.9 &4.8\\
CorrCLIP~\cite{zhang2024corrclip} & \ding{55} &\underline{91.6} &\underline{74.1} &\underline{47.7} &\underline{45.5} &\underline{40.3} &43.6 &\underline{30.6} &- &- &-\\
\rowcolor[HTML]{EFEFEF} 
\textbf{\name~w/ SAM (Ours)} & \ding{55}  &\textbf{93.2} &\textbf{82.2} &\textbf{59.0} &\textbf{53.1} &\textbf{44.6} &\textbf{48.2} &\textbf{33.3} &\textbf{28.2} &\textbf{15.8} &\textbf{8.8}\\
\bottomrule[0.8pt]
\end{tabular}%
}
\caption{\textbf{Comparison to state-of-the-art training-free OVS approaches.} The best results are \textbf{bolded}, with the second-best \underline{underlined}. 
We also analyze data robustness by varying the image resources from the default COCO-2017 to VOC and ADE, respectively (w/o SAM version), where leading performances over SAM-free baselines are \textcolor{gray}{\textbf{bolded}}.}
\label{tab:mIoU_compare_main}
\end{table*}

\subsection{Experimental Setup}
\label{subsec:exp_setup}

\noindent\textbf{Datasets for standard benchmarks.}
We conducted experiments on 10 widely-used OVS benchmarks to evaluate \name, including validation splits of Pascal VOC (VOC)~\cite{everingham2010pascal}, Pascal Context (PC)~\cite{mottaghi2014role}, COCO Object (Object)~\cite{lin2014microsoft}, COCO Stuff (Stuff)~\cite{caesar2018coco}, Cityscapes (City)~\cite{cordts2016cityscapes}, and ADE20K~\cite{zhou2019semantic}. Specifically, VOC has 20 object classes (VOC-20). PC involves PC-459 with 459 classes and PC-59 with 59 frequent classes. Object and Stuff provide COCO-2017 image annotations of 81 and 171 classes, respectively. City captures urban street scenes with 19 classes. ADE20K includes A-150 and A-849 with 150 and 847 classes. For VOC-20 and PC-59, we consider pixels not belonging to any class as \textit{``background''}, represented by VOC-21 and PC-60, respectively. We use the standard mean Intersection-over-Union (mIoU) to measure OVS performance.

\noindent\textbf{Implementation.}
We obtain textual descriptions of input images with LLaVA-1.5~\cite{liu2023llava}.
For class-agnostic segmenter, we deploy Felzenszwalb’s algorithm by default, which is a lightweight superpixel-based algorithm, following the same settings as~\cite{barsellotti2024training}.
The initial pairing is performed by CLIP~\cite{radford2021learning} with ViT-B. 
In the data-enhancing and retrieval phases, we use CLIP-encoded text features, and employ DINOv2 with ViT-L as the default visual feature encoder. Hyperparameters \(\delta_{filter}\), \(k_{sim}\) are both set to \(30\). To compare with SAM-involved OVS baselines, we optionally use SAM as the segmenter, where we prompt a \(32 \times 32\) point grid to obtain masks. More implementation details are in supplementary.

Our reference set construction requires only images as input---no GT captions, classes, or masks, and we default to using COCO-2017~\cite{lin2014microsoft,caesar2018coco} images, which depict everyday scenes with objects in their natural contexts.

\subsection{Comparison to the state-of-the-arts}
\label{subsec:comparison}

\noindent\textbf{Baselines.}
We compare \name~with 14 training-free OVS approaches, including
ReCo~\cite{shin2022reco}, MaskCLIP~\cite{zhou2022extract}, SCLIP~\cite{wang2024sclip}, NACLIP~\cite{hajimiri2024pay}, GEM~\cite{bousselham2024grounding}, PnP~\cite{luo2024emergent}, FreeDA~\cite{barsellotti2024training}, RIM~\cite{wang2024image}, OVDiff~\cite{karazija2024}, CLIPtrase~\cite{shao2024explore}, DiffSegmenter~\cite{wang2023diffusion}, CaR~\cite{sun2024clip}, ProxyCLIP~\cite{lan2024proxyclip}, and CorrCLIP~\cite{zhang2024corrclip}.
Since post-processing techniques for mask refinements, such as DenseCRF~\cite{krahenbuhl2011efficient}, enhance OVS performance but introduce additional computational overhead, we indicate whether each method applies such refinements. Moreover, as SAM is widely regarded as a strong backbone for segment-related tasks, we conduct our comparison across both SAM-free and SAM-involved approaches.

\noindent\textbf{Comparison.}
Table~\ref{tab:mIoU_compare_main} presents the quantitative comparison results, where we observe that \name~consistently outperforms training-free OVS baselines across ten evaluated benchmarks. Notably, \name~achieves superior performance in challenging benchmarks with complex scenarios and large numbers of classes, including A-150, PC-459, and A-847. On Cityscapes, an increase of 8.4 mIoU points (11.3 mIoU increase for SAM-involved comparison) further indicates that our approach excels in domain-specific scenarios.
We also provide qualitative results of \name~in Fig.~\ref{fig:data_more_qua}.

\noindent\textbf{Data robustness.}
To evaluate the data robustness of \name, we vary image resources of our reference set, replacing the default COCO-2017 images with those from VOC and ADE. As shown in Table~\ref{tab:mIoU_compare_main}, SAM-free \name~maintains strong performance over baselines. Across 10 benchmarks, \name~achieves the highest mIoU in 6 benchmarks with VOC and 7 with ADE. These results highlight the effectiveness and flexibility of our approach in diverse settings.

\subsection{Analysis, Ablation, and Discussion}

\label{subsec:ablation}

\begin{figure}[tb]
  \centering
   \includegraphics[width=0.95\linewidth]{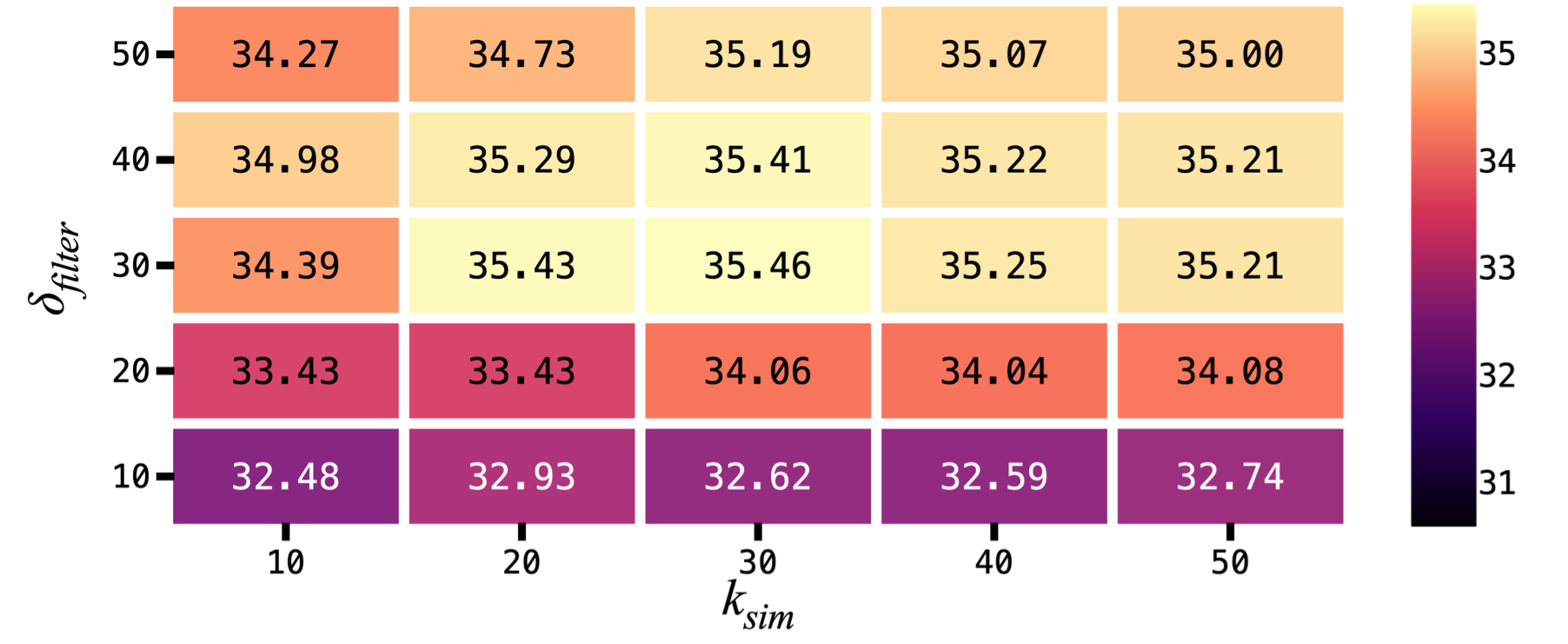}
   \caption{Hyperparameter analysis. \(\delta_{filter}\) is the drop ratio in \textit{group-based filtering}; \(k_{sim}\) is the number of top-similar pairs in \textit{semantic enriching}.}
   \label{fig:hypers_main}
\end{figure}

\noindent\textbf{Hyperparameters.}
Our data-enhancement process involves two key hyperparameters: \textit{the drop ratio per group \((\delta_{filter})\)} in \textit{group-based filtering} (i), and \textit{the number of top-similar label pairs \((k_{sim})\)} in \textit{semantic enriching} (ii). 
To determine their values, we perform a grid search on $1k$ ($1\%$) randomly sampled images from the training split of COCO Stuff (with no overlap with evaluation data), as shown in Fig.~\ref{fig:hypers_main}.
Initially, increasing both parameters improves performance, as more misaligned data is filtered out and label diversity is enhanced. However, beyond a certain point, further increases cause a decline in performance: a high drop ratio removes correct pairings and reduces data diversity, while an excessive \(k_{sim}\) introduces noise by including less-similar labels. We use 30 as the default value for both parameters in our implementation.

\noindent\textbf{Data enhancing component analysis.}
We analyze the effects of our data enhancement components: (i) group-based filtering and (ii) semantic enriching. As shown in Table~\ref{tab:ablation_enhancement_components}, the base set without enhancement yields relatively low performance, highlighting the critical role of data quality. We also observe higher individual effectiveness of (i), while enriching labels alone provides modest gains due to data noises. Notably, the performance improves significantly when both are integrated, demonstrating the complementary benefits of our data enhancement components.

\begin{table}[tb]
\centering
\resizebox{\columnwidth}{!}{%
\begin{tabular}{p{4cm}<{\raggedright} c c c c}
\toprule[0.8pt]
\multirow{2}{*}{Components}
& \multicolumn{4}{c}{mIoU}                                                                                  \\ \cline{2-5}
             & VOC-20 & PC-59 & Object  & A-150 \\ \hline
Base set (no enhancement)         & 70.03 & 35.42       & 39.38       & 22.03    \\
w/ (i) Group-based filtering      & 91.10       & 40.66       &  42.48      & 24.09 \\
w/ (ii) Semantic enriching   & 79.50       &  36.69      & 39.92       &23.41   \\\rowcolor[HTML]{EFEFEF}
w/ (i) and (ii)     &  \textbf{92.34}             &\textbf{44.89}            & \textbf{45.50}       &  \textbf{26.13} \\ \bottomrule[0.8pt]
\end{tabular}%
}
\caption{Impact of data enhancement components.}
\label{tab:ablation_enhancement_components}
\end{table}

\noindent\textbf{Analysis of different data-cleaning approaches.} 
We apply group-based filtering to remove \(\delta_{filter}\%\) of outliers from each label-identified group (refer to Sec.~\ref{subsec:refset_refinement}). This experiment examines the impact of different data-cleaning strategies by comparing three methods: (a) the common approach of dropping segment-label pairs with the globally lowest cross-modal similarity scores (CLIP scores); (b) a variant that also uses CLIP scores, removing segment-label pairs within each constructed groups; (c) our group-based filtering based on intra-modal feature similarity. By only changing this module, we evaluate OVS performance as shown in Table~\ref{tab:ablation_seg_drop}, where we keep the drop ratio fixed at $30\%$. The results indicate that (a) and (b) perform much worse than (c), suggesting that filtering based on CLIP scores is less effective, likely due to its weaker alignment with actual pairing quality.

\begin{table}[tb]
\centering
\resizebox{\columnwidth}{!}{%
\begin{tabular}{p{3.9cm}<{\raggedright} c c c c}
\toprule[0.8pt]
  & \multicolumn{4}{c}{mIoU}                                          \\ \cline{2-5} 
\multirow{-2}{*}{Data cleaning} & VOC-20         & PC-59           & Object          & A-150         \\ 
\toprule[0.5pt]
Global CLIP score$^{(a)}$         & 79.34 & 39.79 & 40.79 & 21.18 \\
Group-based CLIP score$^{(b)}$              & 80.05          & 41.84          & 41.81          & 22.79         \\\rowcolor[HTML]{EFEFEF} 
Ours $^{(c)}$          & \textbf{92.34}          & \textbf{44.89}          & \textbf{45.50}          & \textbf{26.13}        \\
\bottomrule[0.8pt]
\end{tabular}
}
\caption{Analysis of different data filtering approaches.}
\label{tab:ablation_seg_drop}
\end{table}

\begin{table}[tb]
\centering
\resizebox{\columnwidth}{!}{%
\begin{tabular}{p{3cm}<{\centering} p{1.2cm}<{\centering} p{1.2cm}<{\centering} p{1.2cm}<{\centering} p{1.2cm}<{\centering}}
\toprule[0.8pt]
\multirow{2}{*}{Feature encoder}
&\multicolumn{4}{c}{mIoU}                                                                                  \\ \cline{2-5}
             & VOC-20 &PC-59 &Object  & A-150 \\ \toprule[0.5pt]
CLIP$_B$     &  91.61        & 36.51   & 39.82       &  24.72  \\
CLIP$_L$     &  92.16        & 37.63   & 40.52       &  25.56  \\
DINOv$2_B$   & 91.72   & 43.65    & 44.71    & 25.29   \\
\rowcolor[HTML]{EFEFEF}
DINOv$2_L$   & \textbf{92.34}    & \textbf{44.89}   & \textbf{45.50}    & \textbf{26.13}  \\
\bottomrule[0.8pt]
\end{tabular}%
}
\caption{Analysis of visual encoder variations.}
\label{tab:ablation_Feature_main}
\end{table}

\noindent\textbf{Analysis of different visual encoders.}
We investigate the impact of different visual feature encoders, including CLIP and DINOv2 with ViT-B and ViT-L architectures, denoted by CLIP$_B$, CLIP$_L$, DINOv2$_B$, and DINOv2$_L$, respectively. As shown in Table~\ref{tab:ablation_Feature_main}, upgrading from the base to large variants within the same encoder family results in slight performance improvements. When comparing CLIP and DINOv2 at the same scale, DINOv2 achieves better performance as expected, given its specialization in visual representation learning. Notably, by using CLIP alone and removing DINOv2 from our framework, we still achieve performance that is comparable to or better than baselines using similar backbones. This suggests that our approach is robust to variations in visual encoders.

\begin{figure}[t]
  \centering
   \includegraphics[width=1\linewidth]{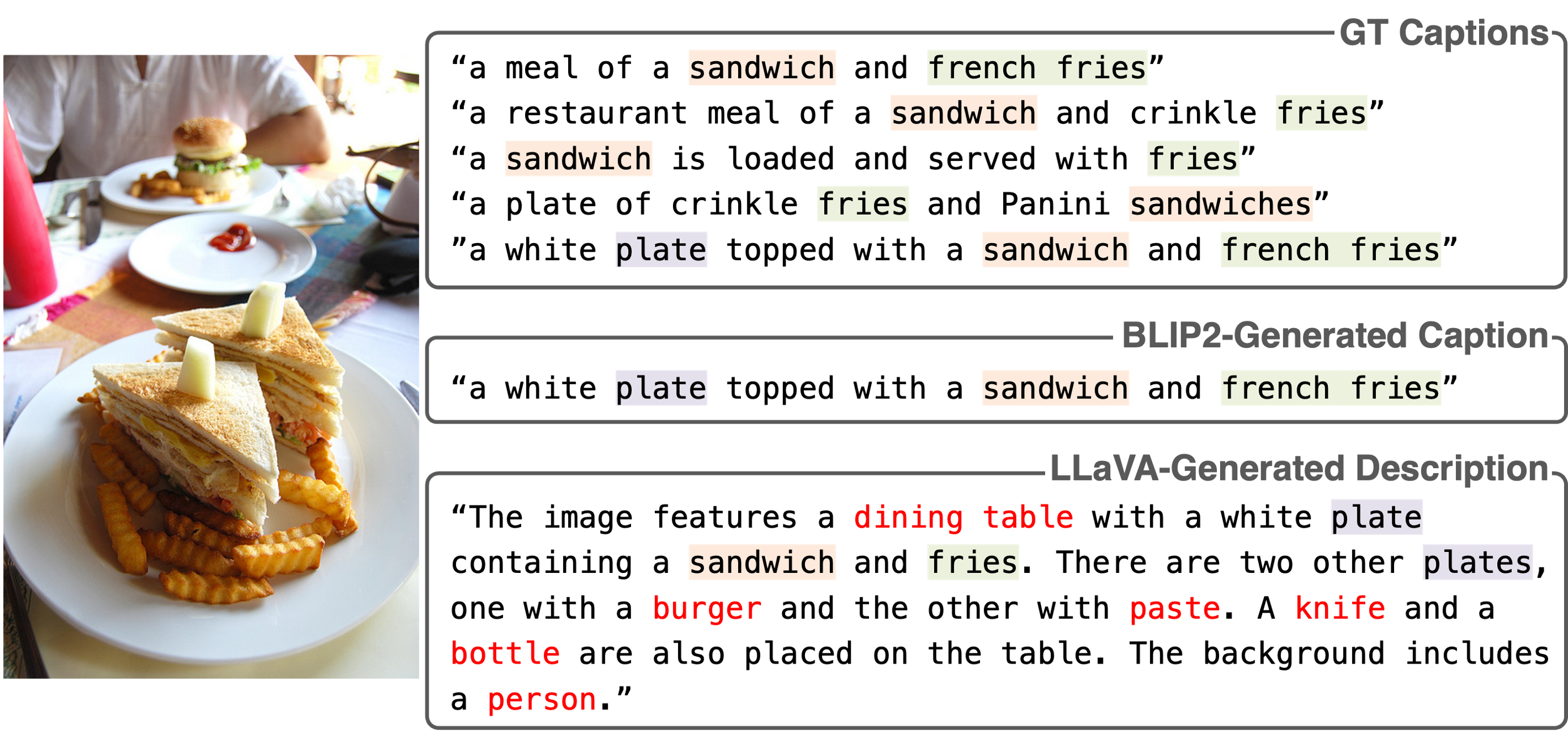}
   \caption{Image descriptions from different resources. {\textcolor{red}{Red text} highlights concepts uniquely present in the LLaVA description.}}
   \label{fig:description_example}
\end{figure}

\begin{table}[t]
\centering
\resizebox{\columnwidth}{!}{%
\begin{tabular}{p{3cm}<{\centering} p{1.2cm}<{\centering} p{1.2cm}<{\centering} p{1.2cm}<{\centering} p{1.2cm}<{\centering}}
\toprule[0.8pt]
\multirow{2}{*}{Description generator}
&\multicolumn{4}{c}{mIoU}                                                                                  \\ \cline{2-5}
             & VOC-20 & PC-59  & Object & A-150 \\ \toprule[0.5pt]
\rowcolor[HTML]{EFEFEF}
LLaVA~\cite{liu2023llava}      &  \textbf{92.34}            & \textbf{44.89}          &   \textbf{45.50}      &  \textbf{26.13}  \\
BLIP-2~\cite{li2023blip}          & 89.41       &   40.85    &      37.64   &  24.31   \\
GT Caption & 89.02 &40.15 &37.77 &24.09  \\
\bottomrule[0.8pt]
\end{tabular}%
}
\caption{Ablation study of the image description generator.}
\label{tab:ablation_Caption_main}
\end{table}

\begin{table}[tb]
\centering
\resizebox{\columnwidth}{!}{%
\begin{tabular}{p{3cm}<{\centering} p{1.2cm}<{\centering} p{1.2cm}<{\centering} p{1.2cm}<{\centering} p{1.2cm}<{\centering}}
\toprule[0.8pt]
\multirow{2}{*}{Segmenter}
&\multicolumn{4}{c}{mIoU}                                                                                  \\ \cline{2-5}
             & VOC-20 &PC-59  &Object  & A-150 \\ \toprule[0.5pt]
\rowcolor[HTML]{EFEFEF}
Superpixel~\cite{felzenszwalb2004efficient}   & 92.34   & 44.89   & 45.50    & 26.13  \\ 
% \rowcolor[HTML]{EFEFEF}
SAM~\cite{kirillov2023segment}          & 93.15   & \textbf{53.10}    & 48.21    & \textbf{28.21}  \\
SAM2~\cite{ravi2024sam2}      &  \textbf{93.18}             & 52.04   & \textbf{48.40}       &  \textbf{28.21}  \\
\bottomrule[0.8pt]
\end{tabular}%
}
\caption{Ablation study of the segmenter.}
\label{tab:ablation_Segmentor_main}
\end{table}

\noindent\textbf{Analysis of the description generator.}
Providing semantically rich image descriptions is important for our data construction. However, not all descriptions are equally informative.
Fig.~\ref{fig:description_example} presents an image from the COCO-2017~\cite{caesar2018coco} dataset alongside its descriptions from three resources: (1) GT captions, (2) BLIP-2~\cite{li2023blip}, and (3) LLaVA~\cite{liu2023llava}, where we highlight the detected visual concepts.
We can observe common concepts presented in all three, ``sandwich'', ``fries'', and ``plate''. However, concepts such as ``dining table'', ``burger'', ``knife'', ``bottle'', and ``person'' are missing from both (1) and (2).
To further study the impact of using BLIP-2 and LLaVA as description generators, we conduct a quantitative evaluation. As shown in Table~\ref{tab:ablation_Caption_main}, LLaVA demonstrates superior performance, achieving significantly higher mIoU compared to BLIP-2, where BLIP-2 captions yield only marginal improvements over GT captions. These results highlight LLaVA's ability to generate richer, more detailed descriptions and quantitatively confirm its effectiveness in enhancing training-free OVS with retrieval.

\noindent\textbf{Analysis of the segmenter.}
We also investigate the impact of different segmenters in Table~\ref{tab:ablation_Segmentor_main} by replacing our default superpixel algorithm-based segmenter with advanced segmentation models, SAM~\cite{kirillov2023segment} and SAM2~\cite{ravi2024sam2}. While a more advanced segmentation model leads to slightly better performance, our data-quality-enhancement pipeline enables even a simple segmentation algorithm to achieve strong results compared to baseline methods.

\noindent\textbf{Analysis of data effectiveness.}
We evaluate data effectiveness by comparing \name~with FreeDA~\cite{barsellotti2024training}, a representative retrieval-based method that also utilizes COCO-2017 dataset~\cite{lin2014microsoft,caesar2018coco}. Unlike our approach, which takes images as input and generates textual descriptions, FreeDA uses GT captions to generate synthetic images for constructing its reference set. As shown in Table~\ref{tab:data_efficiency}, despite FreeDA incorporating five times more input samples, its reference set contains fewer unique labels than \name, due to the limited semantic diversity of GT captions (refer to Fig.~\ref{fig:description_example}). Our data-enhancement strategy improves data quality while reducing dataset size, resulting in a reference set with over a million fewer segment-text pairs than FreeDA. Using this considerably smaller reference set, \name~achieves superior performance even with a much simpler retrieval mechanism, highlighting the effectiveness of high-quality, semantically rich data over quantity.

\begin{table}[t]
\resizebox{\columnwidth}{!}{%
\begin{tabular}{c p{1.1cm}<{\centering}p{1.1cm}<{\centering}p{1.1cm}<{\centering}p{1.1cm}<{\centering}p{1.2cm}<{\centering}}
\toprule[0.8pt] 
\multirow{2}{*}{Methods} & \multicolumn{2}{c}{Input data} & \multicolumn{3}{c}{Reference set} \\ \cmidrule(l){2-3}  \cmidrule(l){4-6}
 & Format           & Size        & \#labels       & \#pairs    &\ Storage   \\\toprule[0.5pt]
\rowcolor[HTML]{EFEFEF} 
Ours                                    & Images           & 118k        & 41k        & 1,023k       & 4GB         \\
FreeDA~\cite{barsellotti2024training}   & Captions         & 591k        & 19k        & 2,167k        & 17GB       \\ 
\bottomrule[0.8pt]
\end{tabular}%
}
\caption{Comparing data effectiveness with FreeDA, a retrieval-based OVS baseline that uses a synthetic reference set.}
\label{tab:data_efficiency}
\end{table}

\begin{table}[t]
\centering
\resizebox{\columnwidth}{!}{%
\begin{tabular}{p{3cm}<{\centering} p{1.2cm}<{\centering} p{1.2cm}<{\centering} p{1.2cm}<{\centering} p{1.2cm}<{\centering}}
\toprule[0.8pt]
Methods  & PC-59 & A-150  & PC-459 & A-847 \\ \toprule[0.5pt]
ProxyCLIP~\cite{lan2024proxyclip}  &0.17&0.27  &0.21   &0.32   \\
SCLIP~\cite{wang2024sclip}      &0.20  & 0.34  &0.57   &0.53  \\
CaR~\cite{sun2024clip} &5.06 &14.67 &33.46 &75.09 \\ \hline
FreeDA~\cite{barsellotti2024training}    & 0.89  &0.89 &4.89  &5.01  \\
\rowcolor[HTML]{EFEFEF}
Ours   &0.29 &0.34 &0.31 &0.34 \\
\bottomrule[0.8pt]
\end{tabular}%
}
\caption{Inference time comparison (seconds/image).}
\label{tab:infertime}
\end{table}

\noindent\textbf{Analysis of inference time.}
Inference time for training-free OVS varies due to factors such as the number of classes, image resolution, and inference strategies. To evaluate this, we compare our method with four representative approaches: (1) ProxyCLIP~\cite{lan2024proxyclip}, which augments CLIP attention with DINO features, (2) SCLIP~\cite{wang2024sclip}, which exploits CLIP self-attention followed by PAMR post-processing~\cite{wannenwetsch2020probabilistic}, (3) CaR~\cite{sun2024clip}, which iteratively queries two CLIP models for mask proposal and classification, and (4) FreeDA~\cite{barsellotti2024training}, a retrieval-based method.
All experiments are conducted on two NVIDIA 4090 GPUs, measuring total inference time per dataset to compute the average. As shown in Table~\ref{tab:infertime}, (3) is the most time-consuming due to its iterative processing. Our method surpasses the retrieval-based baseline (4), benefiting from a smaller reference set and streamlined retrieval design, while remaining competitive with (1) and (2), which do not rely on reference sets.

\section{Conclusion}
\label{sec:conclusion}

In this work, we introduce \name, a data-centric framework for training-free OVS by refining multi-modal embeddings. We observe the overlooked challenge of data quality, and demonstrate its critical impact on this dense scene understanding task. Following our data pipeline, we produce a reference set with well-aligned, rich, and contextually relevant segment-text pairs. Extensive experimental results highlight that enhancing data quality can be more beneficial than relying on complex retrieval mechanisms or model-specific adaptations. We hope this work can inspire future research for exploring data-centric strategies to further improve training-free OVS.

\noindent\textbf{Acknowledgements}
This research is supported in part by the National Institute of Biomedical Imaging and Bioengineering of NIH under Grant No. P41-EB032840.

%%%%%%%%%%%%%%%%%%%%%%%%%%%%%
% {
%     \small
%     \bibliographystyle{ieeenat_fullname}
%     \bibliography{main}
% }

%%%%%%%%%%%%%%%%%%%%%%%%%%%%%
% \clearpage
%%%%%%%%%%%%%%%%%%%%%%%%%%%%%

{
    \small
    \bibliographystyle{ieeenat_fullname}
    \bibliography{main}
}

\appendix

\setcounter{equation}{0}
\renewcommand\theequation{A\arabic{equation}} 
\setcounter{table}{0}  
\setcounter{figure}{0}
\renewcommand{\thetable}{A\arabic{table}}
\renewcommand{\thefigure}{A\arabic{figure}}
\renewcommand{\thealgorithm}{A\arabic{algorithm}}
\clearpage
\setcounter{page}{1}
\maketitlesupplementary
\noindent Our supplementary material is organized as follows:
\begin{itemize}
    \item Sec.~\ref{appendix:definition} provides a formal problem definition of open-vocabulary segmentation (OVS).
    \item Sec.~\ref{appendix:approach} provides additional details of our approach and implementation.
    \item Sec.~\ref{appendix:exp} includes more experimental results and discussions to supplement Sec. 4 of the main paper.
     \begin{itemize}
         \item Sec.~\ref{app_sub:ablation_res}-Additional Ablation Study Results
         \item Sec.~\ref{app_sub:qualitative_res}-Additional Qualitative Results
         \item Sec.~\ref{app_sub:back}-Backbone Usage for Training-Free Methods
         \item Sec.~\ref{app_sub:free}-Free-form Queries and In-the-wild Results
         \item Sec.~\ref{app_sub:datause}-Data Usage for Training-Required Methods 
         \item Sec.~\ref{app_sub:compare_training}-Comparison with Training-required Methods
     \end{itemize}
    \item Sec.~\ref{appendix:future} discusses the limitation of our work.
\end{itemize}

\section{Problem Definition}
\label{appendix:definition}
To complement the definition of open-vocabulary segmentation (OVS), we formulate it mathematically as follows.

Given an input image \( \bm{I} \) and a candidate set of class labels \( \bm{\mathcal{L}} = \{\mathcal{L}_n \}_{n=1}^{N} \), the objective of OVS is to assign a class label \( \mathcal{L}_n \in \bm{\mathcal{L}} \) to each pixel in \( \bm{I} \). Each \( \mathcal{L}_n \) represents the \( n\)-th class described by free-form text, where \( N \) denotes the total number of candidate classes. Unlike traditional semantic segmentation, where the category set is fixed and predefined during training (\( \bm{\mathcal{L}} = \bm{\mathcal{L}}_{\text{train}} \)), OVS allows for segmentation of arbitrary and unseen categories, operating under a zero-shot setting. This flexibility facilitates adaptive and robust dense scene understanding in dynamic real-world scenarios.

\section{Approach and Implementation}
\label{appendix:approach}

\subsection{Prompt for Generating Image Descriptions}
\label{app_sub:prompt}
We design a specific prompt to obtain semantically enriched image descriptions using LLaVA. Our prompt is:

\noindent\textit{``Describe this image in detail. Mention all visible objects, their parts, contexts, and characteristics like size, color, and texture. Also, describe the background/foreground context, including any natural scene or man-made structures, such as wall, ceiling, sky, and cloud. FOCUS ONLY on visible objects or contexts. Avoid speculation or guesses.''}

\subsection{Filtering Ambiguous Labels}
\label{app_sub:enhance_processing}

\noindent\textbf{Object Hallucination in MLLM Outputs.}
Multi-modal large language models (MLLMs) such as LLaVA often suffer from object hallucination. This includes generating descriptions of tangible objects not present in the input image, which we address through our group-based filtering phase. Additionally, MLLMs frequently produce ambiguous outputs reflecting abstract or subjective concepts evoked by the image. For instance, descriptions like ``\textit{The room has a cozy atmosphere}''
\begin{figure}[t!]
  \centering
   \includegraphics[width=1\linewidth]{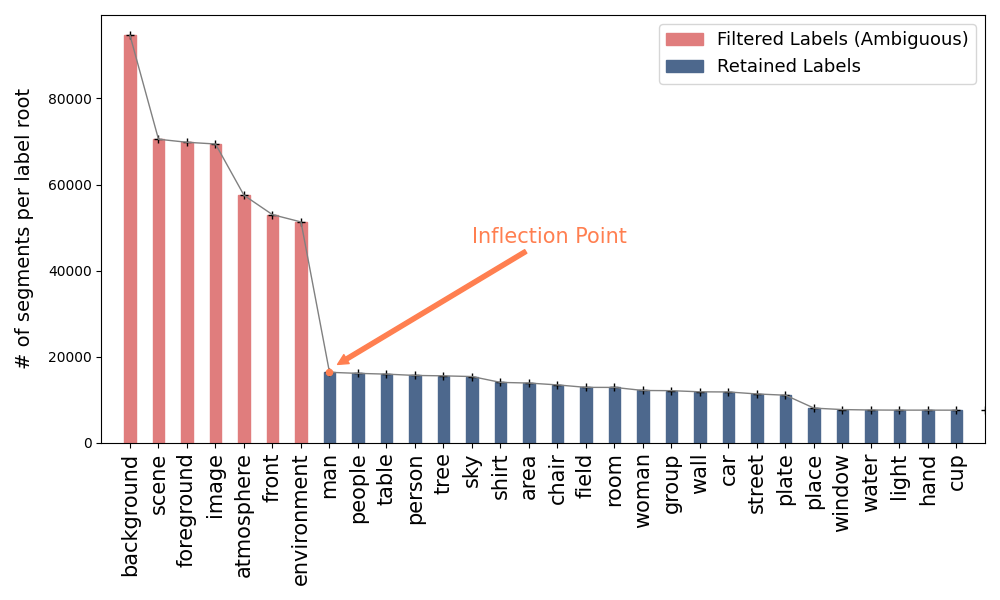}
\caption{\textbf{The number of corresponding segments for each unique label root.} The knee of the distribution curve, \textit{Inflection Point}, indicates the threshold for filtering out ambiguous labels.}
   \label{fig:label_freq}
\end{figure}
\begin{figure}[t!]
  \centering
   \includegraphics[width=1\linewidth]{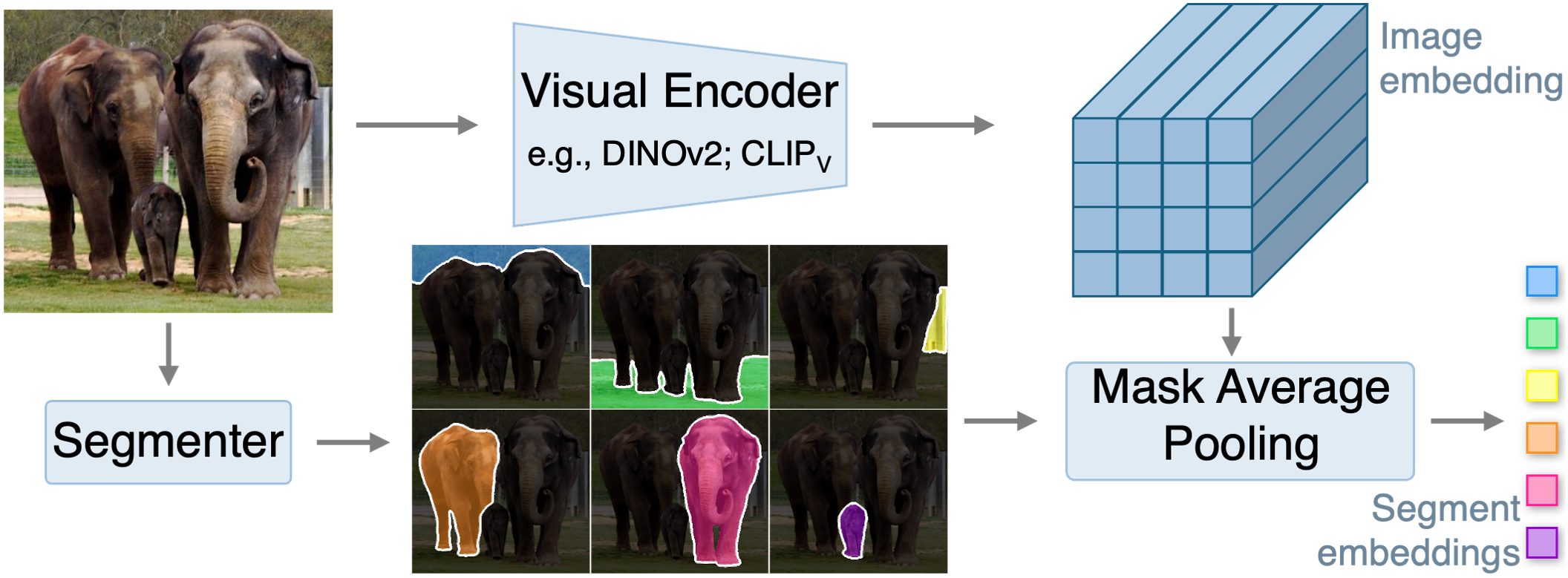}
\caption{Visual feature encoding for image segments.}
   \label{fig:visual_encoder}
\end{figure}
lead to ambiguous labels such as ``\textit{atmosphere},'' which are ungrounded in observable entities and irrelevant to segmentation tasks.

\noindent\textbf{Fast Filtering of Ambiguous Labels.}
To address this issue, we propose a fast and effective approach to eliminate ambiguous labels arising from evoked descriptions. Due to their abstract nature, these labels appear frequently across MLLM-generated descriptions, often corresponding to an unusually large number of segments in the dataset. This observation forms the basis of our aggregation-based analysis. As described in Sec. 3.2, we group segment-text pairs by consistent label roots. For each group represented by a unique label root, we compute the total number of corresponding segments (i.e., the group size) and plot the distribution of group sizes.
\begin{figure*}[th]
  \centering
   \includegraphics[width=1\linewidth]{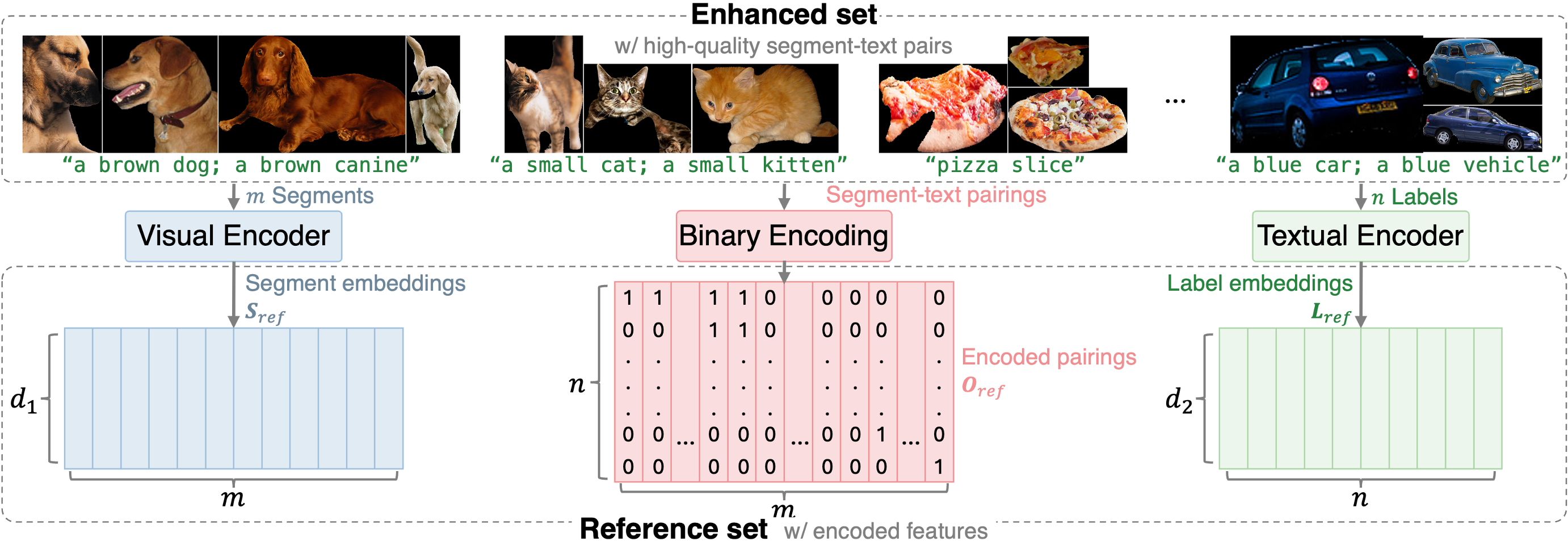}
\caption{\textbf{Illustration of reference set encoding for similarity-based retrieval.} Image segments, textual labels, and their relationships are encoded as \( \bm{S}_{\text{ref}} \), \( \bm{L}_{\text{ref}} \), and \( \bm{O}_{\text{ref}} \), respectively. These embeddings collectively form the reference set, enabling efficient retrieval.}
   \label{fig:reference_set_encoding}
\end{figure*}
By identifying the knee of the curve---referred to as the inflection point (see Fig.~\ref{fig:label_freq})---we filter out labels exceeding this point, such as ``background,'' ``scene,'' ``image,'' and ``atmosphere.'' These labels dominate the dataset and detract from meaningful segmentation labels. Removing them ensures the dataset remains focused on concrete and observable objects, improving its relevance and usability for segmentation tasks.

\subsection{Feature Encoding}
\label{app_sub:feature_encoding}

\noindent\textbf{Visual feature encoding.}
We compute segment embeddings following common practices~\cite{barsellotti2024training,wang2024image,karazija2024,wang2024use}. According to the requirements, a visual encoder such as DINOv2 or CLIP$_V$ is used, denoted as \( \varphi \). As shown in Fig.~\ref{fig:visual_encoder}, given an input image \(\bm{I}\) and its \(K\) corresponding segment masks \(\mathcal{M}={\{\bm{M}_{k}\}}_{k=1}^K\), the visual encoder processes the image to obtain its embedding. To align the segment masks with the encoder's output resolution, the masks are resized using a downscaling function, \(\zeta\). Lastly, we apply mask average pooling (MAP) to produce embedding for each segment \(\bm{S}_k\). This process is represented as:
\begin{equation}
    \bm{S}_k=\text{MAP}(\varphi(\bm{I}), \zeta(\bm{M}_{k})).
\end{equation}

\noindent\textbf{Textual feature encoding.}
We generate text embeddings using a textual encoder CLIP$_T$, denoted by \(\phi\). For a given label \(\mathcal{L}\), we deploy four templates to prompt the encoder: \textit{``A photo of \{\},''} \textit{``This is a photo of \{\},''} \textit{``There is \{\} in the scene,''} and \textit{``A photo of \{\} in the scene.''} The text encoder processes each prompted input, and the resulting embeddings are averaged to form the final label embedding \(\bm{L}\). This process is expressed as:
\begin{equation}
    \bm{L} = \frac{1}{P} \sum_{p=1}^P \phi({\psi}_p(\mathcal{L})),
\end{equation}
where \(P\) is the number of templates, and \({\psi}_p(\mathcal{L})\) represents applying \(p\)-th template to label \(\mathcal{L}\).

All encoded features, regardless of modality, are L2-normalized to facilitate our cosine similarity computation.

\subsection{Reference Set Construction}
\label{app_sub:ref_set}

Following our intra-modality data enhancement phase (refer to Sec. 3.2), we have obtained a high-quality set of segment-text pairs. Fig.~\ref{fig:reference_set_encoding} depicts how we obtain specific embeddings to construct the reference set for streamlined retrieval. 
The visual encoder processes image segments to extract \(d_1\)-dimensional segment embeddings (\( \bm{S}_{\text{ref}} \)), while a textual encoder generates \(d_2\)-dimensional label embeddings (\( \bm{L}_{\text{ref}} \)). To represent the relationships between segments and their associated labels, we utilize binary encoding to formulate \(\bm{O}_{\text{ref}} \in \mathbb{R}^{m \times n}\), where \(m\) and \(n\) are the numbers of unique segments and labels, respectively. Each row of \( \bm{O}_{\text{ref}} \) corresponds to a segment, and a column entry of `1' indicates an association with a specific label and `0' otherwise. 
The resulting reference set is defined by \( \{ \bm{S}_{\text{ref}}, \bm{O}_{\text{ref}}, \bm{L}_{\text{ref}} \} \), combining visual, textual, and relational encodings. This structured representation enables efficient similarity-based retrieval in the subsequent phase.

\subsection{Pseudocode for Similarity-Based Retrieval}
\label{app_sub:retrieval}

To complement Sec. 3.3, we provide a Python-style pseudocode in Alg.~\ref{algo:pseudo} to detail the similarity-based retrieval process. The variable names are consistent with those in Sec. 3.3 for ease of reference, and comments within the pseudocode indicate the steps corresponding to the equations discussed in the main paper.

\begin{algorithm}[t!]
\caption{Pseudocode for similarity-based retrieval}
\label{algo:pseudo}
\includegraphics[width=1\linewidth]{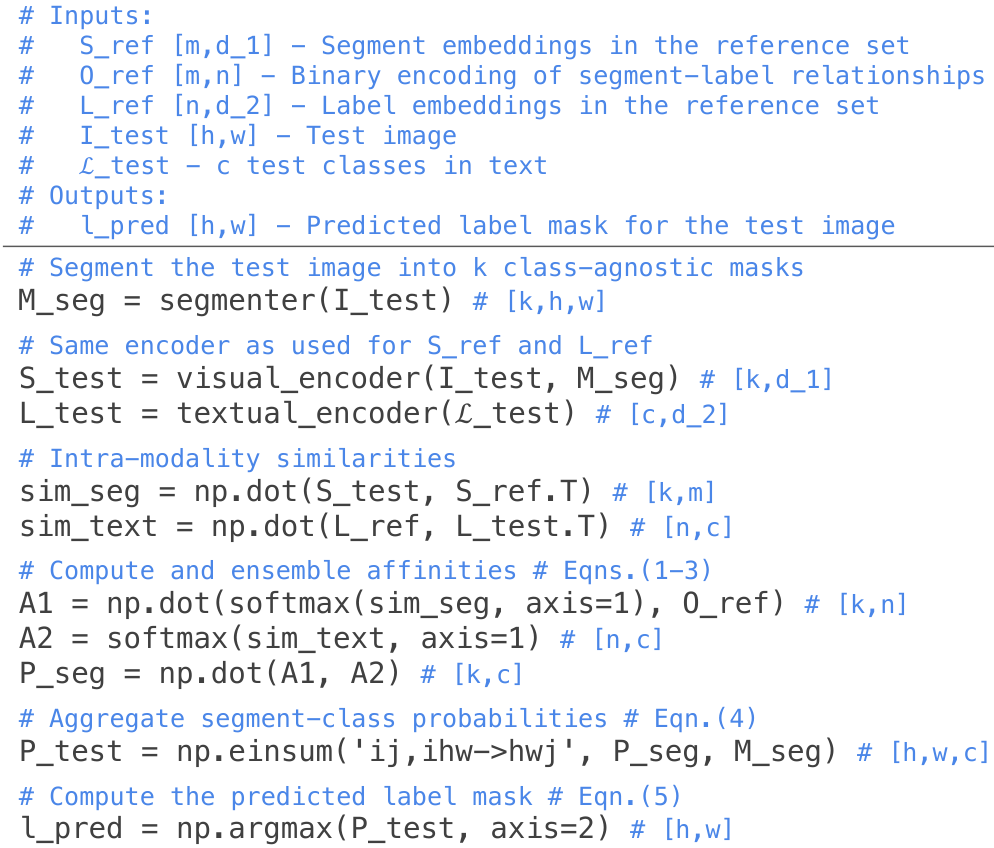}
\end{algorithm}

\section{More Experimental Results and Discussion}
\label{appendix:exp}

\subsection{Additional Ablation Study Results}
\label{app_sub:ablation_res}

In this section, we provide comprehensive results and additional examples to supplement the findings presented in the main paper. The supplementary tables and figures expand on the quantitative and qualitative analyses in Sec. 4.3, offering a more complete view of our ablation studies. We cross-reference the corresponding tables/figures in the main paper for clarity and context.

\begin{figure*}[t]
  \centering
   \includegraphics[width=1\linewidth]{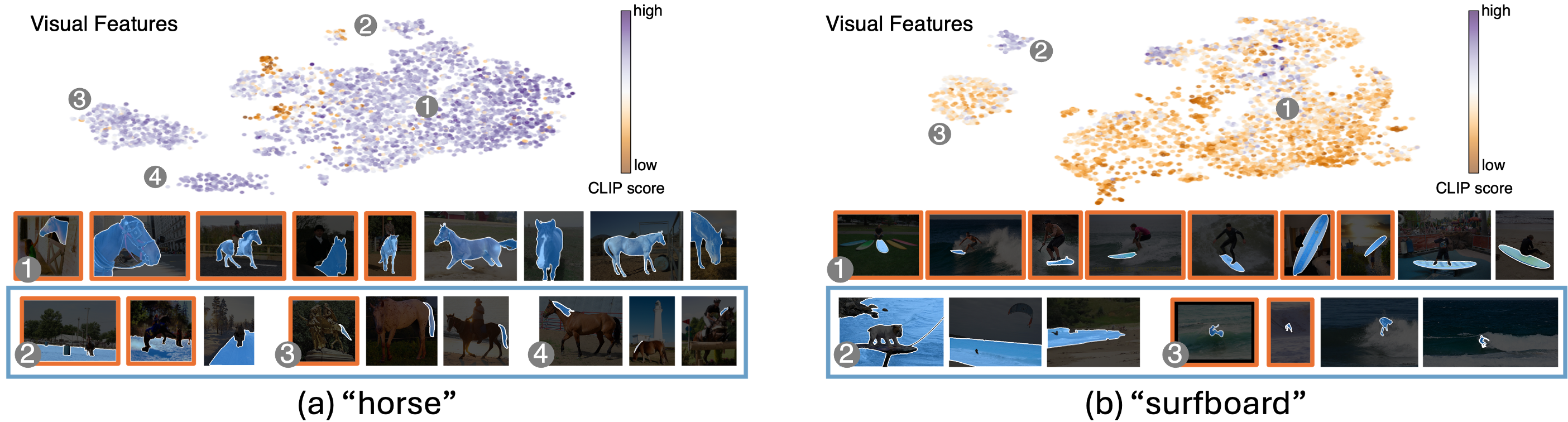}
   \caption{\textbf{The superiority of intra-modality over cross-modality for data issue detection.} Each figure provides a UMAP projection of segment embeddings labeled as ``horse'' or ``surfboard'', respectively, colored by cross-modal similarity scores (CLIP scores) between the segment and its corresponding label. Individual segments are shown below. {\textcolor[HTML]{759DC2}{\textbf{Blue boxes}}} highlight misalignments detected by our filtering; {\textcolor[HTML]{D97843}{\textbf{orange boxes}}} are those detected by low CLIP scores, which remove correct pairings while leaving many misalignments unaddressed.}
   \label{fig:appx_encoders}
\end{figure*}

\begin{figure*}[t]
  \centering
   \includegraphics[width=1\linewidth]{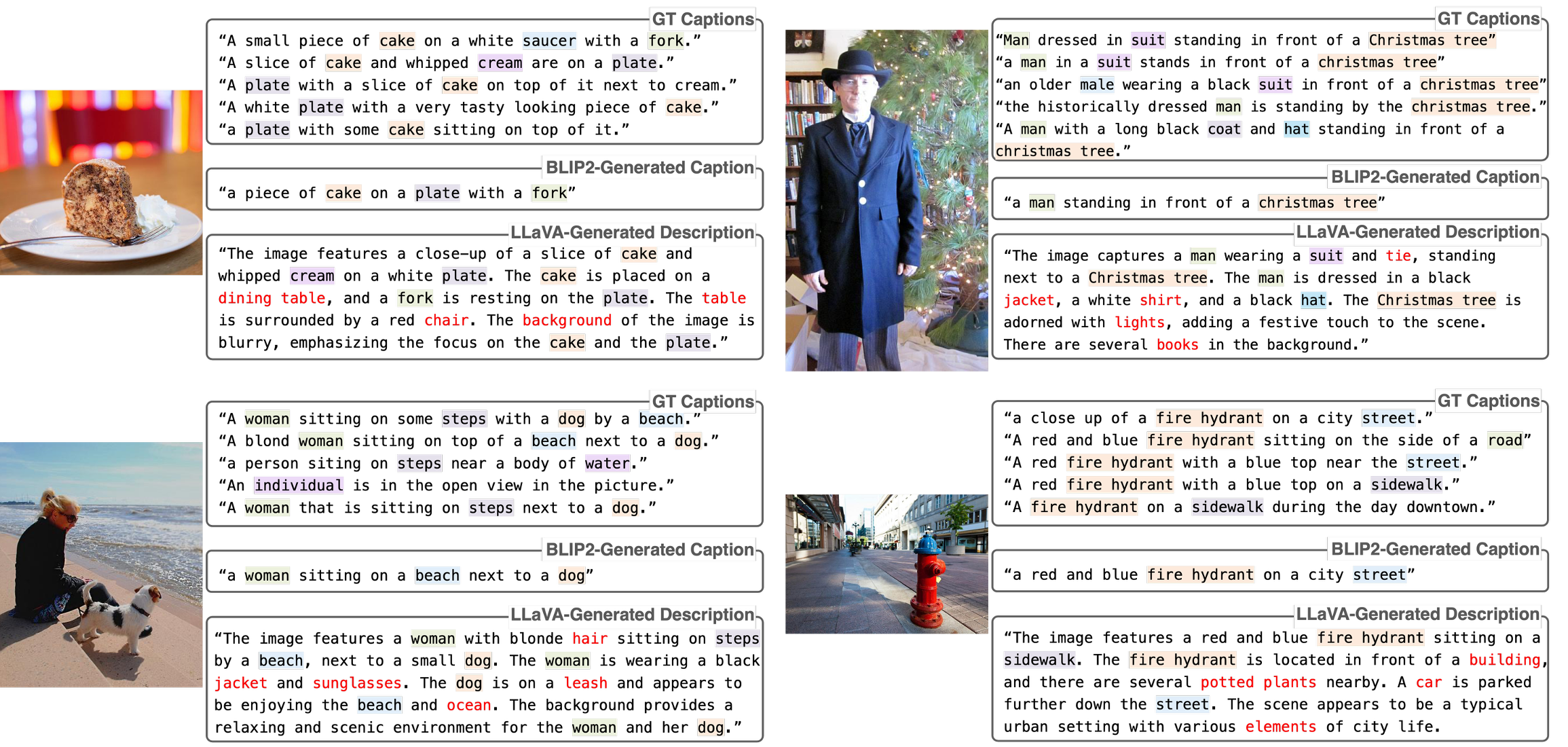}
   \caption{Image descriptions from different resources. {\textcolor{red}{Red text} highlights concepts uniquely present in the LLaVA description.}}
   \label{fig:appx_description_example}
\end{figure*}

\begin{itemize}
    \item \textbf{Data enhancement component analysis.} Full quantitative results for analyzing contributions of individual components in our data enhancement pipeline are presented in Table~\ref{tab:sup_ablation_enhancement_components} (supplementing Table 2 in the main paper).
    \item \textbf{Analysis of different data filtering approaches.} A comprehensive comparison of different data filtering approaches is provided in Table~\ref{tab:appx_ablation_seg_drop}, extending the analysis from Table 3 in the main paper. We include a variant of our group-based filtering, noted as (d). Compared to our default approach that use the same drop ratio for all groups, (d) adapts each group’s drop ratio to its segment consistency, ranging from $0$ to $50\%$, with weights \(w = \frac{1}{n} \sum_{i=1}^n (1 - \langle S_i, S_{center} \rangle)\), allowing for more drops in sparser groups. We can observe that this variant brings further performance gain.
    
    Additionally, we provide more examples to showcase the superiority of intra-modality over cross-modality in Fig.~\ref{fig:appx_encoders}, to complement Fig. 3 in the main paper.
    \item \textbf{Feature encoder backbones.} Full results of using different feature encoder backbones are detailed in Table~\ref{tab:appx_ablation_Feature_main} (bottom), supplementing Table 4 in the main paper.
    \item \textbf{Analysis of the description generator.} Full results on the impact of the description generator are shown in Table~\ref{tab:appx_ablation_Caption_main}, supplementing Table 5. In addition, to further evidence the semantic richness of LLaVA-generated descriptions as discussed in Fig. 7 in the main paper, we provide more examples in Fig.~\ref{fig:appx_description_example}.
    \item \textbf{Analysis of the segmenter.} Additional results for the impact of various segmenters are presented in Table~\ref{tab:appx_ablation_Segmentor_main}, which complements Table 6 in the main paper.
\begin{table}[t!]
\centering
\resizebox{1\columnwidth}{!}{%
\begin{tabular}
{lcccc}
\toprule[0.8pt]
Method & VOC-20& PC-59& A-150 &PC-459   \\ \hline
LLaVA~\cite{liu2023llava} as Classifier &72.65	&35.50	&20.03	&7.22\\
Qwen~\cite{bai2025qwen2} as Classifier &70.67	&36.03	&21.21	&6.36\\
LLaVA~\cite{liu2023llava} as Filter &73.06	&37.55	&22.05	&7.60\\
Qwen~\cite{bai2025qwen2} as Filter &72.18	&38.10	&22.81	&8.15\\\hline
\name  &92.34 &44.89  &26.13 &14.12 \\ 
\name~(OpenFlamingo~\cite{awadalla2023openflamingo}) &92.54	&44.77	&25.95	&13.39  \\
\bottomrule[0.8pt]
\end{tabular}%
}
\caption{\textbf{Top}: Analysis of large MLLM capabilities. We use LLaVA-1.5~\cite{liu2023llava} and Qwen-2.5 VL~\cite{bai2025qwen2} to (1) classify segmentation masks without any references, and (2) perform label filtering for data enhancing, respectively. They all perform significantly worse than~\name. This demonstrates that challenging tasks such as OVS require strategic adaptation rather than direct use. \textbf{Bottom}: Analysis of our performance gain from inherent segment-text pretraining. We replace LLaVA-1.5~\cite{liu2023llava} with OpenFlamingo~\cite{alayrac2022flamingo} trained purely on image-text data. The performance remains comparable, indicating~\name's effectiveness without dense annotations.}
\label{tab:ablation_fm}
\end{table}
\begin{table}[t!]
\centering
\resizebox{1\columnwidth}{!}{%
\begin{tabular}
{lcccc}
\toprule[0.8pt]
Method ($\dag$ w/ seg-text training) & VOC-20& PC-59& A-150 &PC-459   \\ \hline
 CAT-Seg (GT COCO)~\cite{cho2024cat}$\dag$ &94.57 &57.45 &31.81 &19.04\\
 CAT-Seg (\name)$\dag$ &94.60	&59.76	&32.24	&22.03\\
 FreeDA~\cite{barsellotti2024training} &87.91  &43.49  &22.43 &10.24   \\
 FreeDA (\name) &92.35	&44.80	&24.91	&13.89 \\\hline
 \name  &92.34 &44.89  &26.13 &14.12 \\ 
\bottomrule[0.8pt]
\end{tabular}%
}
\caption{\textbf{Data transferability.} We apply \name~data to two representative methods by replacing their training/reference data: (1) training-based CAT-Seg~\cite{cho2024cat}, and (2) retrieval-based FreeDA~\cite{barsellotti2024training}. The results demonstrate the strong utility of our data across both training-based and training-free OVS.}
\label{tab:ablation_data_trans}
\end{table}
    \item \textbf{Analysis of the large MLLM capabilities.} To analyze the capabilities of large MLLM compared to our data enhancement framework, we perform two experiments. (1) We directly leverage advanced MLLMs, including LLaVA-1.5~\cite{liu2023llava} and Qwen-2.5 VL~\cite{bai2025qwen2}, to assign class labels to class-agnostic segmentation masks, without using any data as references. (2) We perform data filtering with each MLLM, rather than using our group-based data filtering. The results are shown in Table~\ref{tab:ablation_fm}, marked as ``* as Classifier'', and ``* as Filter'', respectively. They perform significantly worse than ReME.  This observation aligns with widely discussed challenges in directly using VLMs for fine-grained data matching---they tend to hallucinate object labels and produce noisy predictions. These results highlight: while pre-trained models present potential, challenging tasks like reasoning segmentation~\cite{lai2024lisa} or OVS require strategic adaptation rather than direct use. For instance, LISA~\cite{lai2024lisa} fine-tunes vLLM+SAM backbones, while \name~studies data-centricity---they contribute in complementary ways.
    \item \textbf{Data transferability.}
    We apply \name~data to two representative methods by replacing their training/reference data: (1) training-based CAT-Seg~\cite{cho2024cat}, and (2) retrieval-based FreeDA~\cite{barsellotti2024training}. As shown in Table~\ref{tab:ablation_data_trans}, CAT-Seg (\name) even surpasses the version trained on COCO ground-truth, and FreeDA (\name) also outperforms the original version with its default reference set.
    The results demonstrate the strong utility of \name~data across both training-based and training-free OVS settings.
\end{itemize}

\subsection{Additional Qualitative Results}
\label{app_sub:qualitative_res}
\begin{figure*}[th]
  \centering
   \includegraphics[width=1\linewidth]{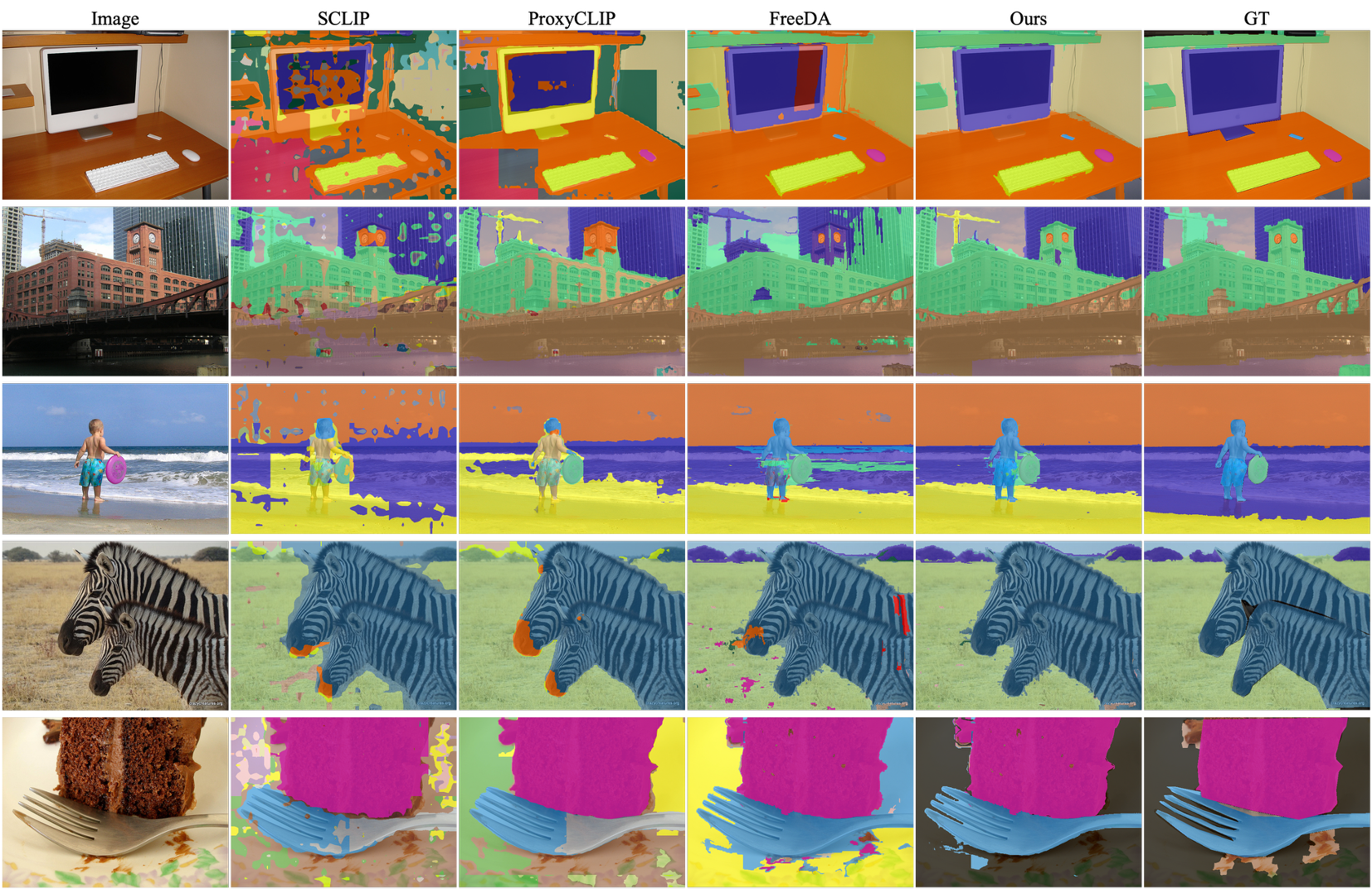}
   \caption{\textbf{Qualitative results of \name~in comparison with other training-free OVS methods.} SCLIP is based on CLIP attention; ProxyCLIP enhances CLIP attention with DINO features; FreeDA and \name~are retrieval-based methods, adopting the same superpixel-algorithm~\cite{felzenszwalb2004efficient} for class-agnostic segmentation. We observe increasing quality of OVS results from left to right, with less noise in both masks and assigned labels.}
   % Best viewed in color with zoom-in.
   \label{fig:appx_pixel_compare}
\end{figure*}

We perform additional qualitative comparisons with other training-free baselines. The results are shown in Fig.~\ref{fig:appx_pixel_compare}.
In addition, we present qualitative results of \name-SAM on datasets with a large number of categories. Specifically, we include ADE20K~\cite{zhou2019semantic} with 847 categories (Fig.~\ref{fig:qual_ADE_847}), Pascal Context~\cite{mottaghi2014role} with 459 categories (Fig.~\ref{fig:qual_pc_459}), and COCO Stuff~\cite{caesar2018coco} with 171 categories (Fig.~\ref{fig:qual_coco_stuff}).

\subsection{Backbone Usage for Training-Free Methods}
\label{app_sub:back}

Table~\ref{tab:backbone_compare} presents the backbone usage across various training-free methods. As shown, earlier approaches predominantly relied on a single CLIP backbone, but their overall performance falls short compared to more recent methods that leverage multiple backbones. 
Compared to these multi-backbone methods, our approach (1) remains entirely off-the-shelf, avoiding structural modifications to the backbone as implemented in ProxyCLIP, and (2) achieves the best performance while maintaining controlled backbone usage.

Additionally, existing methods employ different backbone variants, such as ViT-B/16 and ViT-L/14, with some supporting even larger models like ViT-H/14. In our comparisons, we use ViT-L/14 by default. However, if a method performs better with ViT-B/16, we report the superior result.

\subsection{Free-form Queries and In-the-wild Results}
\label{app_sub:free}

\begin{figure*}[th]
  \centering
   \includegraphics[width=1\linewidth]{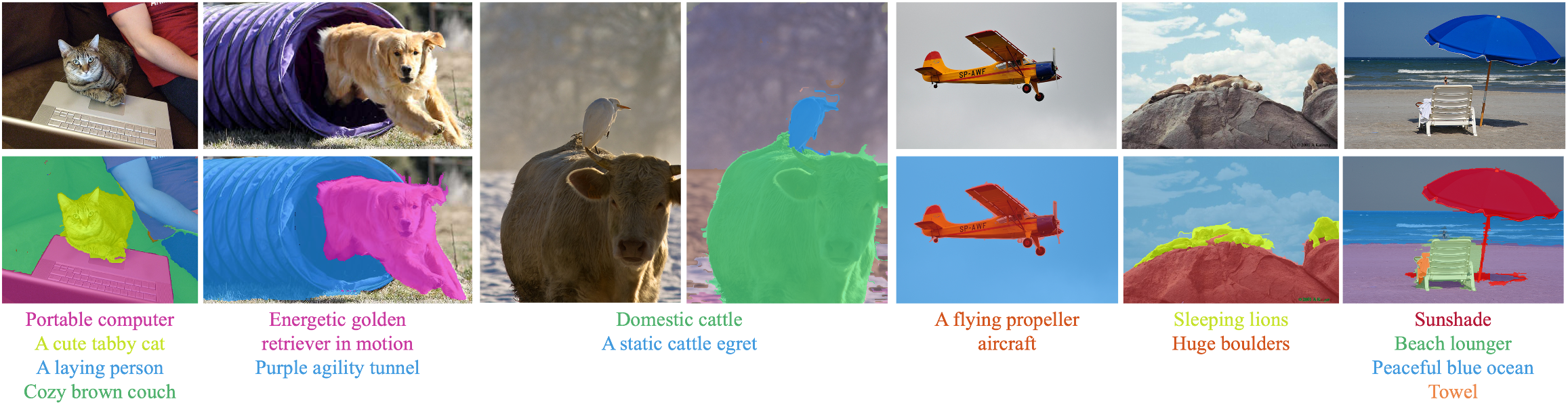}
   \caption{\textbf{In-the-wild segmentation results obtained by prompting \name~with diverse free-form textual inputs.}}
   % Best viewed in color with zoom-in.
   \label{fig:appx_free_form}
\end{figure*}

\begin{figure*}[t]
  \centering
   \includegraphics[width=\linewidth]{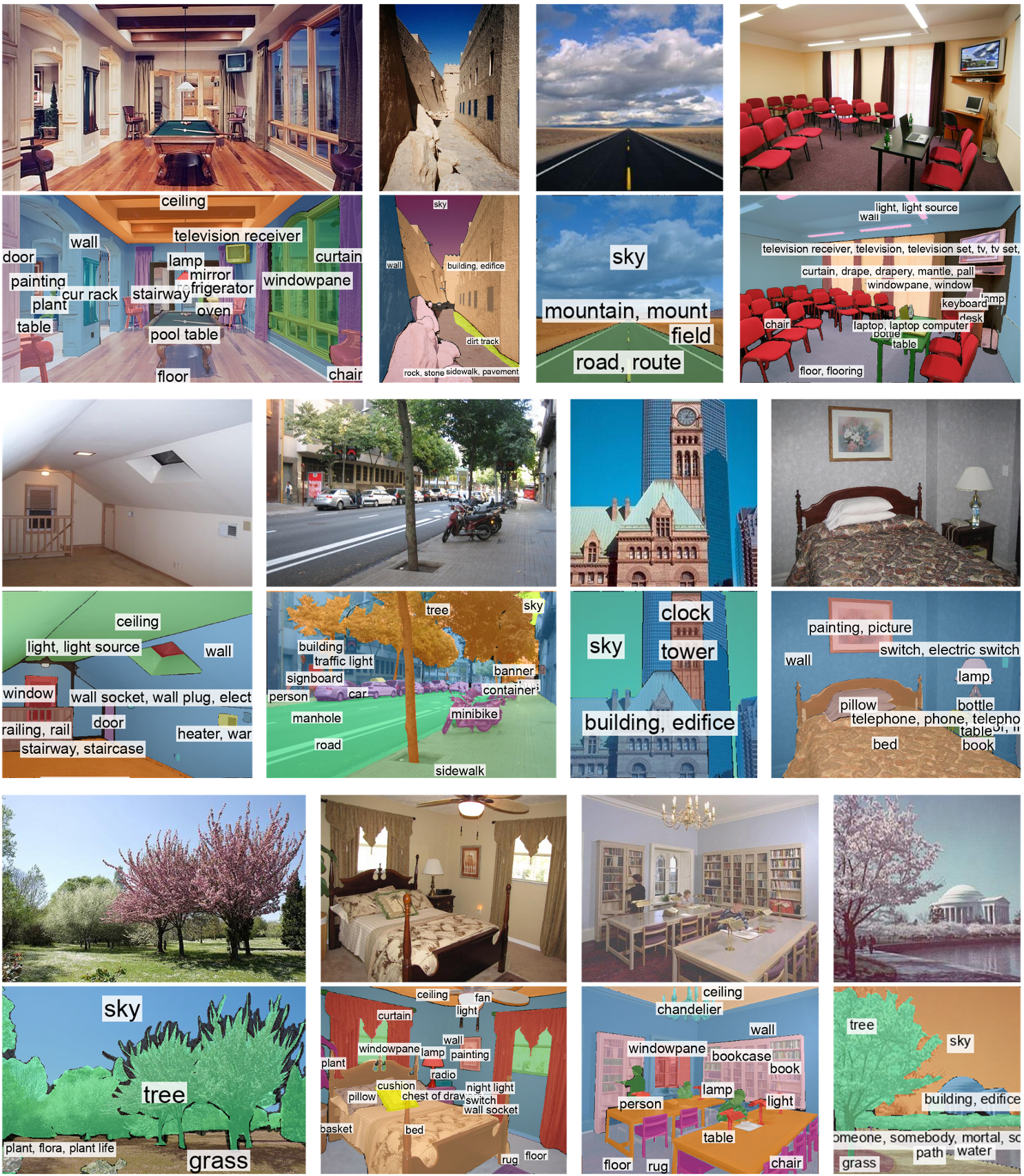}
\caption{\textbf{Qualitative results on ADE20K~\cite{zhou2019semantic} with 847 categories.} }
   \label{fig:qual_ADE_847}
\end{figure*}

\begin{figure*}[t]
  \centering
   \includegraphics[width=\linewidth]{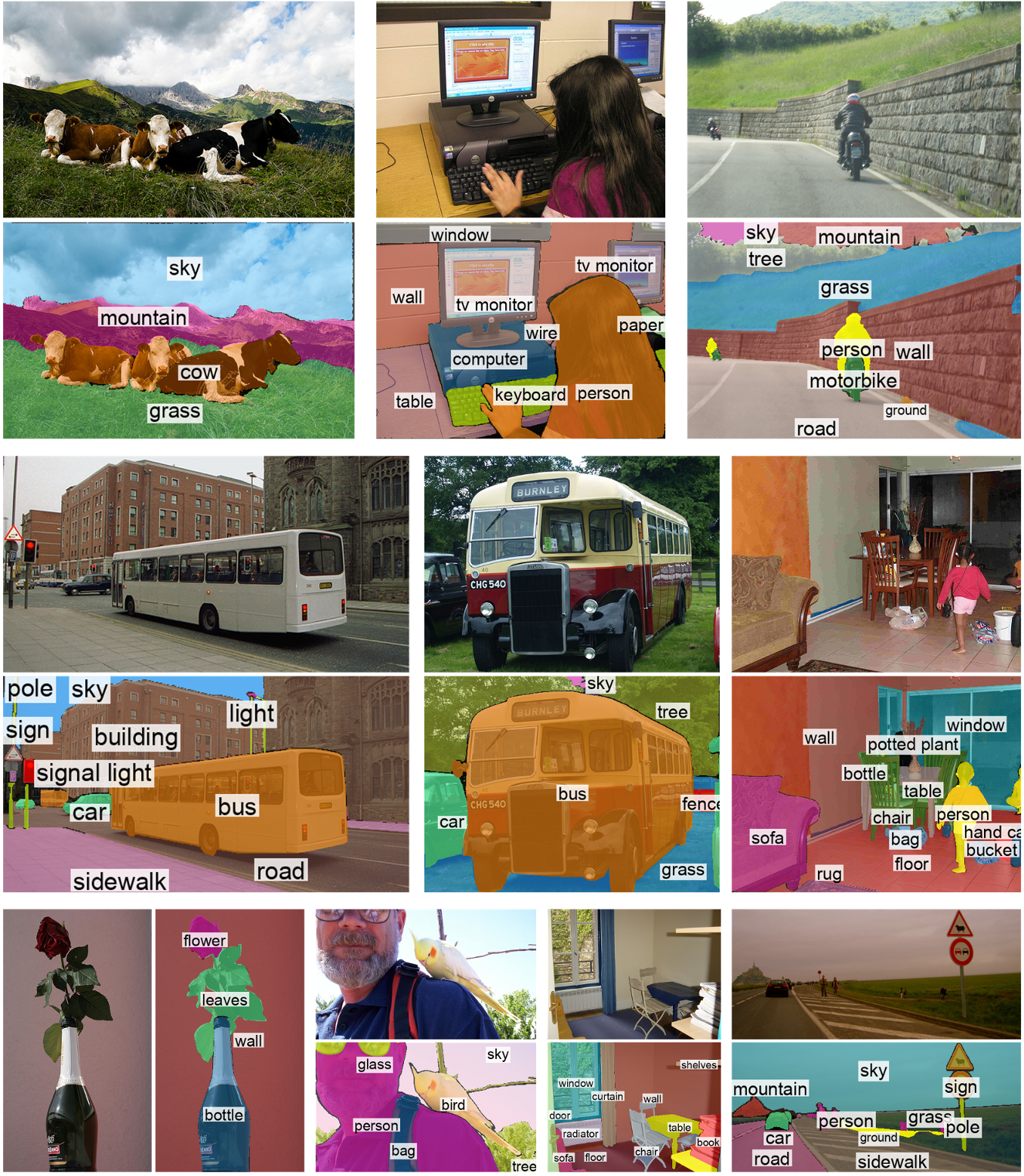}
\caption{\textbf{Qualitative results on Pascal Context~\cite{mottaghi2014role} with 459 categories.} }
   \label{fig:qual_pc_459}
\end{figure*}

\begin{figure*}[t]
  \centering
   \includegraphics[width=\linewidth]{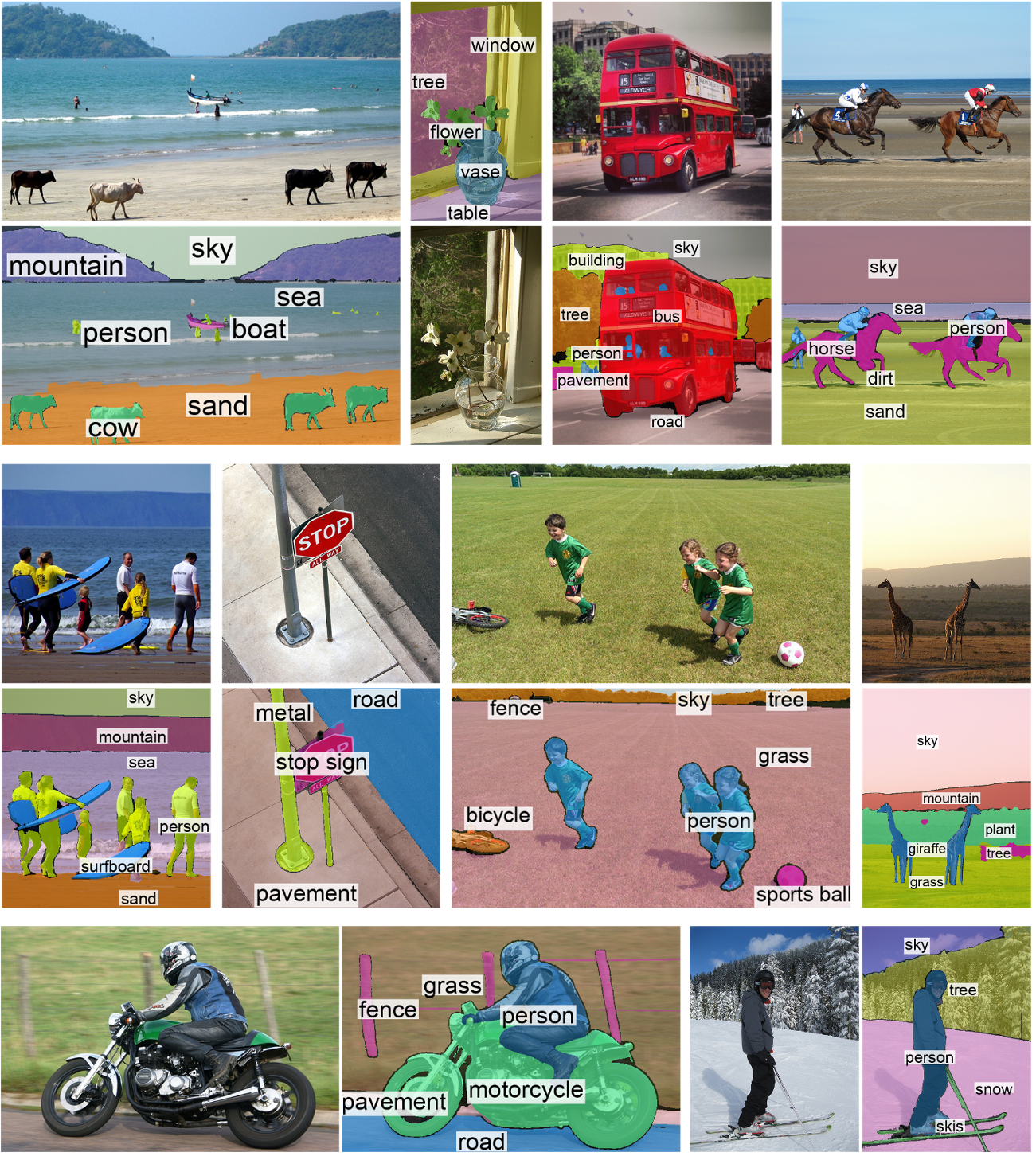}
\caption{\textbf{Qualitative results on COCO Stuff~\cite{caesar2018coco} with 171 categories.} }
   \label{fig:qual_coco_stuff}
\end{figure*}

\noindent\textbf{Generalizability evaluation.}
\textbf{Quantitative.} We evaluate generalizability using free-form text. 
To ensure a fair comparison, we use the same superpixel segmenter as FreeDA.
We prompt GPT4o three times independently to generate diverse free-form class variations (e.g.,``cat''$\to$``small domestic feline'') and then perform retrieval. Results across three runs are summarized in Table~\ref{tab:freeform_main}. 
Shifting to free-from text, FreeDA and ProxyCLIP experience significant performance drops, whereas \name~consistently outperforms them.
\textbf{Qualitative.} Following FreeDA, we collect in-the-wild text and qualitatively evaluate out method. The results are shown in Fig.\ref{fig:appx_free_form}.

\begin{table}[t]
\centering
\resizebox{1\columnwidth}{!}{%
\begin{tabular}{ccc}
\toprule[0.8pt]
Methods & Training or Fine-tuning dataset & Size \\ \hline
GroupViT\cite{xu2022groupvit}    & CC12M+YFCC & 26 million       \\
SimSeg\cite{yi2023simple}      & CC15M & 15 million       \\
TCL\cite{cha2023learning}                   & CC15M & 15 million       \\
CoCu\cite{xing2024rewrite}                  & CC15M+YFCC & 29 million       \\
ZeroSeg\cite{chen2023exploring}                & CC3M+COCO & 3.4 million      \\
OVSegmentor\cite{xu2023learning}           & CC4M & 4 million        \\
SegCLIP\cite{luo2023segclip}               & CC3M+COCO & 3.4 million      \\
CoDe\cite{wu2024image}                   & CC15M & 15 million       \\
SAM-CLIP\cite{wang2024sam}             & CC15M+YFCC+IN21k & 41 million      \\ \bottomrule[0.8pt]
\end{tabular}%
}

\caption{Data usage for training-required OVS methods.}
\label{tab:baseline_data_usage}
\end{table}

\subsection{Data Usage for Training-Required Methods}
\label{app_sub:datause}

For training-required OVS methods using image-text pairs, they often demand extensive training. Table~\ref{tab:baseline_data_usage} provides the training data size for such methods, where we can observe that millions of image-text pairs from diverse datasets are leveraged, indicating their higher computational cost.

\subsection{Comparison with Training-required Methods }
\label{app_sub:compare_training}

Although it falls beyond our primary scope of comparison, we also evaluate our approach against training-required methods, as shown in Table~\ref{tab:mIoU_compare_appx}. 
Our method \textbf{outperforms all approaches fine-tuned with image-text data}. 
When compared to methods fine-tuned with segment-text, our approach surpasses LSeg+~\cite{ghiasi2022scaling}, ZegFormer~\cite{ding2022decoupling}, and ZSseg~\cite{xu2022simple}, but falls short compared to OVSeg~\cite{liang2023open}, SAN~\cite{xu2023side}, and CATSeg~\cite{cho2024cat}. 
This performance gap is commonly observed across all training-free methods when compared to models that demand fine-tuning on segment-text.

However, it is important to note that training-free methods have significantly fewer resources: (1) no training is performed, and (2) no labor-intensive pixel-level annotations. i.e., segment-text data, are required. As a training-free method, we achieve the smallest performance gap compared to these segment-text fine-tuned models.

To sum up, our contributions remain distinct:
\textbf{A.} \textbf{\name~achieves state-of-the-art performance among all training-free methods while also surpassing models trained on millions of image-text pairs}, demonstrating reduced dependence on large-scale training.  
\textbf{B.} Our framework provides a novel perspective on multi-modal data quality, offering contributions that extend beyond OVS.

\section{Limitation}
\label{appendix:future}

One limitation of our framework is the decision to drop misaligned pairs in the base set rather than correcting them by reassigning appropriate labels. For instance, in Fig. 3 of the main paper, misaligned pairs where ``dog'' is associated with segments not depicting dogs are simply filtered out. A more sophisticated approach could involve identifying the correct segments for those labels and reassigning appropriate labels to the affected segments. This refinement would increase the diversity of the final reference set and further enhance the quality of the resulting segment-text embeddings. However, given the diversity and scale of our image resource, COCO-2017~\cite{caesar2018coco}, we opt for a simpler and more efficient data enhancement phase.

In domains with limited data availability and constrained diversity~\cite{malik2025towards}, this limitation could be addressed easily through a plug-in component. After group-based filtering, this component could leverage intra-modality similarity to identify the closest neighbors for each element in misaligned pairs, enabling the estimation of correct matches with minimal computational overhead.

\begin{table*}[t]
\centering
\resizebox{\textwidth}{!}{%
\begin{tabular}{l c c c c c c c c c c c}
\toprule[0.8pt]
\multirow{2}{*}{Components}
& \multicolumn{11}{c}{mIoU}                                                                                  \\ \cline{2-12}
             & VOC-20 & VOC-21 & City & PC-59 & PC-60 & Object & Stuff & A-150      & PC-459 & A-847  & AVG$^{10}$ \\ \hline
Base set (no enhancement)        &70.03	&62.30	&30.94	&35.42	&30.46	&39.38	&27.01	&22.03	&9.14	&6.19  &33.29 \\
w/ (i) Synonym-guided enriching   &79.50	&66.91	&33.47	&36.69	&34.81	&39.92	&28.02	&23.41	&9.56	&6.22	&35.85\\
w/ (ii) Group-based filtering      &91.10	&76.41	&47.36	&40.66	&38.52	&42.48	&31.80	&24.09	&12.96	&7.13	&41.25   \\ \rowcolor[HTML]{EFEFEF}
w/ Both (i) and (ii)     &\textbf{92.34}	&\textbf{79.63}	&\textbf{50.42}	&\textbf{44.89}	&\textbf{41.64}	&\textbf{45.50}	&\textbf{33.12}	&\textbf{26.13}	&\textbf{14.12}	&\textbf{8.43}	&\textbf{43.62}\\ \bottomrule[0.8pt]
\end{tabular}%
}
\caption{Impact of data enhancement components.}
\label{tab:sup_ablation_enhancement_components}
\end{table*}

\begin{table*}[t]
\centering
\resizebox{\textwidth}{!}{%
\begin{tabular}{l c c c c c c c c c c c}
\toprule[0.8pt]
  & \multicolumn{11}{c}{mIoU}                                          \\ \cline{2-12} 
\multirow{-2}{*}{Data filtering alternatives} & VOC-20 & VOC-21 & City & PC-59 & PC-60 & Object & Stuff & A-150   & PC-459  & A-847 & AVG$^{10}$         \\ 
\toprule[0.5pt]
Global filtering*$^{(a)}$         &79.34	&71.19	&41.37	&39.79	&37.25	&40.79	&31.16	&21.18	&11.71	&7.88	&38.47\\
\hspace{0.5em}*Group-based filtering (with cross-modality CLIP score)$^{(b)}$             & 80.05	&72.92	&43.06	&41.84	&39.44	&41.81	&31.88	&22.79	&12.19	&8.33      &39.63    \\
\rowcolor[HTML]{EFEFEF} 
\hspace{0.5em}*Group-based filtering (with intra-modality similarity score)$^{(c)}$          &\textbf{92.34}	&\textbf{79.63}	&\textbf{50.42}	&44.89	&41.64	&45.50	&33.12	&26.13	&14.12	&8.43	&43.62         \\
\hspace{0.5em}*Group-based filtering (with intra-modality similarity score; weighted ratio)$^{(d)}$   & 92.26	&79.61	&50.38	&\textbf{44.97}	&\textbf{41.88}	&\textbf{45.60}	&\textbf{33.17}	&\textbf{26.51}	&\textbf{14.74}	&\textbf{8.58}	&\textbf{43.77}        \\ 
\bottomrule[0.8pt]
\end{tabular}
}
\caption{Analysis of different data filtering approaches.}
\label{tab:appx_ablation_seg_drop}
\end{table*}

\begin{table*}[t]
\centering
\resizebox{\textwidth}{!}{%
\begin{tabular}{p{3cm}<{\centering} c c c c c c c c c c c}
\toprule[0.8pt]
  & \multicolumn{11}{c}{mIoU}                                          \\ \cline{2-12} 
\multirow{-2}{*}{Feature encoder} & VOC-20 & VOC-21 & City & PC-59 & PC-60 & Object & Stuff & A-150  & PC-459 & A-847 & AVG$^{10}$         \\ 
\toprule[0.5pt]
CLIP        &91.61	&68.77	&38.53	&36.51	&35.08	&39.82	&26.85	&24.72	&13.76	&7.51	&37.81 \\
{DINOv2}$_B$             & 91.72	&79.13	&50.20	&43.65	&41.37	&44.71	&32.58	&25.29	&13.79	&7.68	&43.01         \\
\rowcolor[HTML]{EFEFEF} 
{DINOv2}$_L$          &\textbf{92.34}	&\textbf{79.63}	&\textbf{50.42}	&\textbf{44.89}	&\textbf{41.64}	&\textbf{45.50}	&\textbf{33.12}	&\textbf{26.13}	&\textbf{14.12}	&\textbf{8.43}	&\textbf{43.62}        \\
\bottomrule[0.8pt]
\end{tabular}
}
\caption{Analysis of feature encoder variations.}
\label{tab:appx_ablation_Feature_main}
\end{table*}

\begin{table*}[t]
\centering
\resizebox{\textwidth}{!}{%
\begin{tabular}{c c c c c c c c c c c c}
\toprule[0.8pt]
\multirow{2}{*}{Captioners}
& \multicolumn{11}{c}{mIoU}                                                                                  \\ \cline{2-12}
           & VOC-20 & VOC-21 & City & PC-59 & PC-60 & Object & Stuff & A-150      & PC-459 & A-847  & AVG$^{10}$ \\ \hline
             \rowcolor[HTML]{EFEFEF}
LLaVA~\cite{liu2023llava}     &\textbf{92.34}	&\textbf{79.63}	&\textbf{50.42}	&\textbf{44.89}	&\textbf{41.64}	&\textbf{45.50}	&\textbf{33.12}	&\textbf{26.13}	&\textbf{14.12}	&\textbf{8.43}	&\textbf{43.62}  \\
BLIP-2~\cite{li2023blip}        & 89.41	& 56.32	& 40.06	& 40.85	& 38.42	& 37.64	& 30.76	& 24.31	& 12.42	& 6.47	&37.67     \\ 
GT Caption  &89.02	&55.57	&40.19	&40.15	&38.37	&37.77	&29.68	&24.09	&11.64	&5.37	&37.18 \\\bottomrule[0.8pt]
\end{tabular}%
}
\caption{Ablation study of the image description generator.}
\label{tab:appx_ablation_Caption_main}
\end{table*}

\begin{table*}[t]
\centering
\resizebox{\textwidth}{!}{%
\begin{tabular}{c c c c c c c c c c c c}
\toprule[0.8pt]
\multirow{2}{*}{Segmenters}
& \multicolumn{11}{c}{mIoU}                                                                                  \\ \cline{2-12}
           & VOC-20 & VOC-21 & City & PC-59 & PC-60 & Object & Stuff & A-150      & A-847 & PC-459 & AVG$^{10}$ \\ \hline
             \rowcolor[HTML]{EFEFEF}
Superpixel~\cite{felzenszwalb2004efficient}       & 92.34	&79.63	&50.42	&44.89	&41.64	&45.50	&33.12	&26.13	&14.12	&8.43	&43.62     \\ 
SAM~\cite{kirillov2023segment}        & 93.15       & 82.20       & 59.04       & \textbf{53.10} &\textbf{44.58} &48.21 &33.32 &28.21 &15.82 &8.80  & 46.64      \\
SAM2~\cite{ravi2024sam2}    &  \textbf{93.18}             &\textbf{82.26}           & \textbf{61.19}       &  52.03 &43.42 &\textbf{48.40} &\textbf{33.36} &\textbf{28.21}  &\textbf{8.83} &\textbf{15.97} &\textbf{46.69}  \\

\bottomrule[0.8pt]
\end{tabular}%
}
\caption{Ablation study of the segmenter.}
\label{tab:appx_ablation_Segmentor_main}
\end{table*}

\begin{table*}[t]
\centering
\resizebox{\textwidth}{!}{%
\begin{tabular}{c c c c| c c c| c c c| c c c}
\toprule[0.8pt]
\multirow{2}{*}{Methods}
&\multicolumn{12}{c}{mIoU}                                                                                  \\ \cline{2-13}
          & PC-59  & PC-59$^*$  &$\Delta (\%)$ & A-150 & A-150$^*$  &$\Delta (\%)$ & PC-459 & PC-459$^*$  &$\Delta (\%)$ & A-847 & A-847$^*$  &$\Delta (\%)$ \\ \toprule[0.5pt]
\rowcolor[HTML]{EFEFEF}
Ours     &  \textbf{44.89}            & \textbf{42.89{\footnotesize{$\pm$0.9}}}          &    \textbf{$\downarrow$ 4.46}      &  \textbf{26.13}  &  \textbf{26.12{\footnotesize{$\pm$0.3}}} &    \textbf{$\downarrow$ 0.04} &  \textbf{14.12}  &  \textbf{13.14{\footnotesize{$\pm$0.1}}} &    \textbf{$\downarrow$ 6.94} &  \textbf{8.43}  &  \textbf{7.35{\footnotesize{$\pm$0.1}}} &   $\downarrow$ 12.81\\
FreeDA~\cite{barsellotti2024training}          & 43.50      &   36.18{\footnotesize{$\pm$0.8}}    &$\downarrow$ 16.83        & 22.4      &   16.27{\footnotesize{$\pm$1.0}}    &$\downarrow$ 27.37   & 10.20      &   7.16{\footnotesize{$\pm$0.2}}    &$\downarrow$ 29.80    & 5.30      &   2.09{\footnotesize{$\pm$0.1}}    &$\downarrow$ 54.52  \\
ProxyCLIP~\cite{lan2024proxyclip}   & 37.7     &   33.15{\footnotesize{$\pm$1.2}}    &$\downarrow$ 12.05 & 22.6    &   17.12{\footnotesize{$\pm$0.3}}    &$\downarrow$ 24.26 & 11.20      &   8.41{\footnotesize{$\pm$0.3}}    &$\downarrow$ 24.84 & 6.70      &   6.39{\footnotesize{$\pm$0.2}}    & \textbf{$\downarrow$ 4.63}      \\
\bottomrule[0.8pt]
\end{tabular}%
}
\caption{Generalizability evaluation with free-form queries.}
\label{tab:freeform_main}
\end{table*}

\begin{table*}[t]
\centering
\resizebox{\textwidth}{!}{%
\begin{tabular}{l c c c c c c c c c c c c}
\toprule[0.8pt]
\multirow{2}{*}{Methods}  & \multirow{2}{*}{Backbone}   & \multirow{2}{*}{Post-proc}       & \multicolumn{10}{c}{mIoU}                                                                                  \\ \cline{4-13} 
                                                  &     &          & VOC-20 & VOC-21 & City & PC-59 & PC-60 & Object & Stuff & A-150      & PC-459  & A-847  \\ \hline
GEM~\cite{bousselham2024grounding}  &CLIP &\ding{55} &46.2 &24.7 &-&32.6 &21.2 &- &15.1 &10.1 &4.6 &3.7\\
MaskCLIP~\cite{zhou2022extract}   & CLIP, DeepLabV2 &  \Checkmark   & 74.9   & 38.8   & 12.6          & 25.5  & 23.6  & 20.6        & 14.6       & 9.8          & -       & -         \\
ReCo~\cite{shin2022reco}           & CLIP, DenseCLIP &  \Checkmark                                       & 62.4   & 27.2   & 23.2       & 24.7  & 21.9  & 17.3        & 16.3       & 12.4         & -       & -       \\
SCLIP~\cite{wang2024sclip} &CLIP & \Checkmark &83.5 &61.7 &34.1 &36.1 &31.5 &32.1 &23.9 &17.8 &9.3 &6.1\\
CaR~\cite{sun2024clip}  & CLIP & \Checkmark          & \underline{91.4}   & 67.6     & 15.1            & 39.5  & 30.5  & 36.6 & 11.2 & 17.7         & 11.5     & 5.0 \\
NACLIP~\cite{hajimiri2024pay} &CLIP &\Checkmark &83.0 &64.1 &38.3 &38.4 &35.0 &36.2 &25.7 &19.1 &9.0 &6.5\\
CLIPtrase~\cite{shao2024explore}    &CLIP &\Checkmark           & 81.2   & 53.0  & 21.1      & 34.9 & 30.8 & \underline{39.6}       & 24.1      & 17.0        & 9.9    & 5.9    \\
PnP~\cite{luo2024emergent}      & CLIP, GPT4om, BLIP &   \Checkmark     & 79.1  & 51.3   & 19.3      & 31.0    & 28.0    & 36.2        & 17.9       & 14.2         & 5.5    & 4.2    \\
FreeDA~\cite{barsellotti2024training}   & CLIP, Stable Diffusion, DINO & \Checkmark  & 87.9   & 55.4   & 36.7       & \underline{43.5}  & \underline{38.3}  & 37.4        &\underline{28.8}       & 22.4         & 10.2  & 5.3 \\
ProxyCLIP~\cite{lan2024proxyclip}        & CLIP, DINO  & \ding{55}  & 83.2   &  60.6    & \underline{40.1}         & 37.7  & 34.5  & 39.2      &  25.6      & \underline{22.6}         &  11.2  & \underline{6.7}   \\

DiffSegmenter~\cite{wang2023diffusion}    &Stable Diffusion, BLIP, U-Net, DeepLabV2  &  \Checkmark      &  71.4     & 60.1   & -         & 27.5  &  25.1     & 37.9        &   -         & -             &  -       & -        \\
OVDiff~\cite{karazija2024}   & CLIP, Stable Diffusion, GPT, CutLER & \Checkmark   & 80.9   & \underline{68.4}   & 23.4       & 32.9  & 31.2  & 36.2        & 20.3       & 14.1         & \underline{12.0}       & 6.6       \\
\rowcolor[HTML]{EFEFEF} 
\textbf{\name~(Ours)} & CLIP, LLaVA, DINO & \ding{55}  & \textbf{92.3}       &\textbf{79.6}      & \textbf{50.4}           & \textbf{44.9}     &\textbf{41.6}        & \textbf{45.5}            & \textbf{33.1}           & \textbf{26.1}             &\textbf{14.1}        & \textbf{8.4} \\
\bottomrule[0.8pt]
\end{tabular}%
}
\caption{\textbf{Comparison to training-free methods without SAM.} The best overall results are \textbf{bolded}, with the second-best results \underline{underlined}.}
\label{tab:backbone_compare}
\end{table*}

\begin{table*}[t]
\centering
\resizebox{\textwidth}{!}{%
\begin{tabular}{l c c c c c c c c c c c}
\toprule[0.8pt]
\multirow{2}{*}{Methods}    & \multirow{2}{*}{Post-processing}       & \multicolumn{10}{c}{mIoU}                                                                                  \\ \cline{3-12} 
                                                  &               & VOC-20 & VOC-21 & City & PC-59 & PC-60 & Object & Stuff & A-150      & PC-459  & A-847   \\
                                                  \hline
\multicolumn{2}{c}{\textcolor{gray}{\textit{Methods that require finetuning on segment-text data}}}                   &        &        &            &       &       &             &            &              &         &         \\
\textcolor{gray}{LSeg+}\cite{ghiasi2022scaling} & \textcolor{gray}{\ding{55}} &\textcolor{gray}{-} &\textcolor{gray}{59.0}      &\textcolor{gray}{-}  &\textcolor{gray}{36.0} &\textcolor{gray}{-} &\textcolor{gray}{-} &\textcolor{gray}{-} &\textcolor{gray}{13.0} &\textcolor{gray}{5.2} &\textcolor{gray}{2.5} \\
\textcolor{gray}{ZegFormer}~\cite{ding2022decoupling} & \textcolor{gray}{\ding{55}} &\textcolor{gray}{86.2} &\textcolor{gray}{62.7}      &\textcolor{gray}{-}  &\textcolor{gray}{42.8} &\textcolor{gray}{-} &\textcolor{gray}{-} &\textcolor{gray}{-} &\textcolor{gray}{16.9} &\textcolor{gray}{9.1} &\textcolor{gray}{4.9} \\
\textcolor{gray}{ZSseg}~\cite{xu2022simple} & \textcolor{gray}{\ding{55}} &\textcolor{gray}{88.4} &\textcolor{gray}{-}      &\textcolor{gray}{-}  &\textcolor{gray}{44.7} &\textcolor{gray}{-} &\textcolor{gray}{-} &\textcolor{gray}{-} &\textcolor{gray}{20.5} &\textcolor{gray}{-} &\textcolor{gray}{7.0} \\
\textcolor{gray}{OVSeg}~\cite{liang2023open} & \textcolor{gray}{\ding{55}} &\textcolor{gray}{94.5} &\textcolor{gray}{-}      &\textcolor{gray}{-}  &\textcolor{gray}{55.7} &\textcolor{gray}{-} &\textcolor{gray}{-} &\textcolor{gray}{-} &\textcolor{gray}{29.6} &\textcolor{gray}{12.4} &\textcolor{gray}{9.0}  \\      \textcolor{gray}{SAN}~\cite{xu2023side} & \textcolor{gray}{\ding{55}} &\textcolor{gray}{94.6} &\textcolor{gray}{-}      &\textcolor{gray}{-}  &\textcolor{gray}{57.7} &\textcolor{gray}{-} &\textcolor{gray}{-} &\textcolor{gray}{-} &\textcolor{gray}{32.1} &\textcolor{gray}{15.7} &\textcolor{gray}{12.4}  \\      \textcolor{gray}{CAT-Seg}~\cite{cho2024cat} & \textcolor{gray}{\ding{55}} &\textcolor{gray}{97.0} &\textcolor{gray}{82.5}      &\textcolor{gray}{-}  &\textcolor{gray}{63.3} &\textcolor{gray}{-} &\textcolor{gray}{-} &\textcolor{gray}{-} &\textcolor{gray}{37.9} &\textcolor{gray}{23.8} &\textcolor{gray}{16.0}                                 \\ \hline
\multicolumn{2}{c}{\textcolor{gray}{\textit{Methods that require finetuning on image-text data}}}                   &        &        &            &       &       &             &            &              &         &         \\
\textcolor{gray}{GroupViT}~\cite{xu2022groupvit}                  & \textcolor{gray}{\ding{55}}                        & \textcolor{gray}{74.1}  & \textcolor{gray}{52.3}   &   \textcolor{gray}{11.1}       & \textcolor{gray}{23.4}  & \textcolor{gray}{22.4}  & \textcolor{gray}{24.3}        & \textcolor{gray}{15.3}       & \textcolor{gray}{10.6 }& \textcolor{gray}{4.9}  & \textcolor{gray}{4.3}         \\
\textcolor{gray}{SimSeg}~\cite{yi2023simple}                       &\textcolor{gray}{ \Checkmark}          &\textcolor{gray}{57.4}   & \textcolor{gray}{ 35.2}    & \textcolor{gray}{10.8}        & \textcolor{gray}{26.2}  &  \textcolor{gray}{23.4}    & \textcolor{gray}{29.7}        & \textcolor{gray}{18.5}         &  \textcolor{gray}{11.4}          &  \textcolor{gray}{5.0} & \textcolor{gray}{4.7}            \\
\textcolor{gray}{TCL}~\cite{cha2023learning}       & \textcolor{gray}{\Checkmark}                 & \textcolor{gray}{83.2}   & \textcolor{gray}{55.0}   & \textcolor{gray}{23.1}       & \textcolor{gray}{33.9}  & \textcolor{gray}{30.4}  & \textcolor{gray}{31.6}        & \textcolor{gray}{19.6}       & \textcolor{gray}{17.1}        & \textcolor{gray}{8.7}  & \textcolor{gray}{6.3}          \\
\textcolor{gray}{CoCu}~\cite{xing2024rewrite}     &  \textcolor{gray}{\ding{55}}          &  \textcolor{gray}{73.0}     & \textcolor{gray}{51.4}   & \textcolor{gray}{22.1}       & \textcolor{gray}{26.5}    & \textcolor{gray}{23.6}  & \textcolor{gray}{22.7}        & \textcolor{gray}{15.2}       & \textcolor{gray}{12.3}         & \textcolor{gray}{5.1}  & \textcolor{gray}{4.5}           \\
\textcolor{gray}{ZeroSeg}~\cite{chen2023exploring}          &    \textcolor{gray}{\ding{55}}  & \textcolor{gray}{-}      & \textcolor{gray}{40.8}   & \textcolor{gray}{-}         & \textcolor{gray}{20.4}  & \textcolor{gray}{-}    & \textcolor{gray}{20.2}        & \textcolor{gray}{-}         & \textcolor{gray}{-}            & \textcolor{gray}{-}        &\textcolor{gray}{-}         \\
\textcolor{gray}{OVSegmentor}~\cite{xu2023learning}               & \textcolor{gray}{\ding{55}}                  & \textcolor{gray}{- }     & \textcolor{gray}{53.8}   &   \textcolor{gray}{-}       & \textcolor{gray}{-}     & \textcolor{gray}{20.4}  & \textcolor{gray}{25.1}        & \textcolor{gray}{-}         & \textcolor{gray}{5.6}          & \textcolor{gray}{-}        & \textcolor{gray}{-}        \\
\textcolor{gray}{SegCLIP}~\cite{luo2023segclip}     & \textcolor{gray}{\ding{55}}              & \textcolor{gray}{-}     & \textcolor{gray}{52.6}   & \textcolor{gray}{-}          &  \textcolor{gray}{-}     & \textcolor{gray}{24.7}  & \textcolor{gray}{26.5}        &  \textcolor{gray}{-}          & \textcolor{gray}{8.7}          & \textcolor{gray}{-}       &  \textcolor{gray}{-}      \\
\textcolor{gray}{CoDe}~\cite{wu2024image}        &  \textcolor{gray}{\Checkmark}   & \textcolor{gray}{57.7}   & \textcolor{gray}{-}       & \textcolor{gray}{28.9}       & \textcolor{gray}{30.5}  & \textcolor{gray}{-}      & \textcolor{gray}{32.3}        & \textcolor{gray}{23.9}       & \textcolor{gray}{17.7}         &\textcolor{gray}{-}         & \textcolor{gray}{-}        \\
\textcolor{gray}{SAM-CLIP}~\cite{wang2024sam}     &    \textcolor{gray}{\ding{55}}           &  \textcolor{gray}{-}      & \textcolor{gray}{60.6}   &\textcolor{gray}{-}            & \textcolor{gray}{-}      & \textcolor{gray}{29.2}  &  \textcolor{gray}{-}   & \textcolor{gray}{31.5}       & \textcolor{gray}{17.1}         & \textcolor{gray}{-}       &  \textcolor{gray}{ -}      \\ \hline
\multicolumn{2}{l}{\textit{Training-free Methods without SAM}}                   &        &        &            &       &       &             &            &              &         &         \\
GEM~\cite{bousselham2024grounding}  &\ding{55} &46.2 &24.7 &-&32.6 &21.2 &- &15.1 &10.1 &4.6 &3.7\\
MaskCLIP~\cite{zhou2022extract}    &  \Checkmark    & 74.9   & 38.8   & 12.6          & 25.5  & 23.6  & 20.6        & 14.6       & 9.8          & -       & -         \\
ReCo~\cite{shin2022reco}           &  \Checkmark                                      & 62.4   & 27.2   & 23.2       & 24.7  & 21.9  & 17.3        & 16.3       & 12.4         & -       & -       \\
SCLIP~\cite{wang2024sclip} & \Checkmark &83.5 &61.7 &34.1 &36.1 &31.5 &32.1 &23.9 &17.8 &9.3 &6.1\\
CaR~\cite{sun2024clip}  & \Checkmark          & \underline{91.4}   & 67.6     & 15.1            & 39.5  & 30.5  & 36.6 & 11.2 & 17.7         & 11.5     & 5.0 \\
NACLIP~\cite{hajimiri2024pay} &\Checkmark &83.0 &64.1 &38.3 &38.4 &35.0 &36.2 &25.7 &19.1 &9.0 &6.5\\
CLIPtrase~\cite{shao2024explore}     &\Checkmark           & 81.2   & 53.0  & 21.1      & 34.9 & 30.8 & \underline{39.6}       & 24.1      & 17.0        & 9.9    & 5.9    \\
PnP~\cite{luo2024emergent}       &   \Checkmark      & 79.1  & 51.3   & 19.3      & 31.0    & 28.0    & 36.2        & 17.9       & 14.2         & 5.5    & 4.2    \\
FreeDA~\cite{barsellotti2024training}     & \Checkmark  & 87.9   & 55.4   & 36.7       & \underline{43.5}  & \underline{38.3}  & 37.4        &\underline{28.8}       & 22.4        & 10.2  & 5.3 \\
ProxyCLIP~\cite{lan2024proxyclip}        & \ding{55}    & 83.2   &  60.6    & \underline{40.1}         & 37.7  & 34.5  & 39.2      &  25.6      & \underline{22.6}         &  11.2  & \underline{6.7}   \\
DiffSegmenter~\cite{wang2023diffusion}      &  \Checkmark      &  71.4     & 60.1   & -         & 27.5  &  25.1     & 37.9        &   -         & -             &  -       & -        \\
OVDiff~\cite{karazija2024}     & \Checkmark   & 80.9   & \underline{68.4}   & 23.4       & 32.9  & 31.2  & 36.2        & 20.3       & 14.1         & \underline{12.0}       & 6.6       \\
\rowcolor[HTML]{EFEFEF} 
\textbf{\name~(Ours)}  & \ding{55}  & \textbf{92.3}       &\textbf{79.6}      & \textbf{50.4}           & \textbf{44.9}     &\textbf{41.6}        & \textbf{45.5}            & \textbf{33.1}           & \textbf{26.1}             &\textbf{14.1}        & \textbf{8.4} \\
\hline
\multicolumn{2}{l}{\textit{Training-free Methods with SAM}}                   &        &        &            &       &       &             &            &              &         &         \\
RIM~\cite{wang2024image}      & \ding{55}    & 77.8   & -      & -          & 34.3  & -     & \underline{44.5}        & -          & -            & -       & -       \\
CaR w/ SAM~\cite{sun2024clip} & \ding{55} &- &70.2 &16.9 &40.5 &31.1 &37.6 &12.4 &17.9 &\underline{11.8}   & 5.7\\
CLIPtrase w/ SAM~\cite{shao2024explore} & \ding{55} &82.3 &57.1 &- &36.4 &32.0 &44.2 &24.8 &17.2 &10.6 &6.0 \\
ProxyCLIP w/ SAM~\cite{lan2024proxyclip}  & \ding{55} &80.4 &59.3 &37.0 &37.0 &33.6 &35.4 &25.0 &\underline{19.1} &6.9 &4.8\\
CorrCLIP~\cite{zhang2024corrclip} & \ding{55} &\underline{91.6} &\underline{74.1} &\underline{47.7} &\underline{45.5} &\underline{40.3} &43.6 &\underline{30.6} &- &- &-\\
\rowcolor[HTML]{EFEFEF} 
\textbf{\name~(Ours)} w/ SAM & \ding{55}  &\textbf{93.2} &\textbf{82.2} &\textbf{59.0} &\textbf{53.1} &\textbf{44.6} &\textbf{48.2} &\textbf{33.3} &\textbf{28.2} &\textbf{15.8} &\textbf{8.8}\\
\bottomrule[0.8pt]
\end{tabular}%
}
\caption{\textbf{Comparison to state-of-the-art OVS approaches.} The best overall results are \textbf{bolded}, with the second-best results \underline{underlined}. 
We also analyze data robustness by varying the image resources of our reference set from the default COCO-2017 to VOC and ADE, respectively, where leading performances over baselines are \textcolor{gray}{\textbf{bolded}}.}
\label{tab:mIoU_compare_appx}
\end{table*}

\clearpage

\end{document}